\pdfoutput=1
\documentclass[11pt]{article}

\usepackage[]{EMNLP2022}

\usepackage{times}
\usepackage{latexsym}

\usepackage[T1]{fontenc}
\usepackage[utf8]{inputenc}

\usepackage{graphicx}
\usepackage{multirow}
\usepackage{verbatim}
\usepackage{lipsum}
\usepackage{amssymb}
\usepackage{amsmath}
\usepackage{booktabs}
\usepackage{url} 
\usepackage{enumitem}
\usepackage{mathtools}
\usepackage{subcaption}
\usepackage{fontawesome} 
\usepackage{contour}
\usepackage{textcomp}
\usepackage{amsmath}
\usepackage{microtype}

\usepackage{microtype}

\newcommand{\Reflect}{\emph{Reflect}}

\title{\emph{Reflect}, Not \emph{Reflex}: Inference-Based Common Ground Improves\\ Dialogue Response Quality}
\author{
Pei Zhou \quad Hyundong Cho \quad Pegah Jandaghi \quad  Dong-Ho	Lee \quad Bill Yuchen Lin  \\ \textbf{Jay Pujara \quad Xiang Ren}\\
Department of Computer Science and Information Sciences Institute\\
University of Southern California\\
\small{\texttt{\{peiz,dongho.lee,yuchen.lin,jpujara,xiangren\}@usc.edu}, \texttt{\{jcho,jandaghi\}@isi.edu}}
}

\date{}

\begin{document}
\maketitle

\begin{abstract}
%%% Focus on more one-to-many nature and argue later that explicit inferences is helpful

% 1. engaging responses; 2. what makes human responses engaging; 3. model capability of doing diverse; 4. current way limitation

% big --> narrow scope and make it convincing; intuitive characteristics (engaging or others)
% Common ground might read confusing to some readers

% How to connect common ground with diverse responses; what are intents and their differences between inferences

% Neural response generation (RG) systems often struggle to generate specific and interesting responses.
Human communication relies on \emph{common ground (CG)}, the mutual knowledge and beliefs shared by participants, to produce coherent and interesting conversations.
In this paper, we demonstrate that current response generation (RG) models produce generic and dull responses in dialogues because they act \emph{reflexively}, failing to explicitly model CG, both due to the lack of CG in training data and the standard RG training procedure.
We introduce \Reflect, a dataset that annotates dialogues with explicit CG (materialized as inferences approximating shared knowledge and beliefs) and solicits 9k diverse human-generated responses each following one common ground.
Using \Reflect, we showcase the limitations of current dialogue data and RG models: less than half of the responses in current data is rated as high quality (sensible, specific, and interesting) and models trained using this data have even lower quality, while most \Reflect~responses are judged high quality.
%we propose automated metrics of contextual differentiation, demonstrating that human-generated dialogues have utterances that respond to contextual information while vanilla RG models produce responses that are less differentiated. 
%Humans communicate by first making implicit inferences about speaker’s emotions and experiences to enrich \emph{common ground}, i.e. mutual knowledge and beliefs, and then responding based on the knowledge.
% Moreover, what makes human responses unique and engaging is that there are multiple ways of extending the common ground, which further leads to different responses.
%humans have different intents leading to different inferences that extend the common ground, which further leads to different responses.
% Current RG systems either do not model common ground explicitly or neglect the multi-dimensionality of common ground.
%To simulate this response generation (RG) process, we construct a dataset where each dialogue has multiple human-annotated common ground, materialized as commonsense inferences expressed in question-answer pairs, each associated with next-turn dialogue responses. 
% To test the hypothesis that whether models can generate more human-like responses through human-like communication process, we construct a dataset with explicit and multiple common ground, materialized as commonsense inferences expressed in question-answer pairs, each associated with next-turn dialogue responses. 
% We materialize common ground with potential inferences a dialogue participant makes expressed in question-answer pairs.
% \xiang{Below needs revision.}
Next, we analyze whether CG can help models produce better quality responses by using \Reflect~CG to guide RG models. Surprisingly, we find that simply prompting GPT3 to ``\emph{think}'' about CG generates 30\% more quality responses, showing promising benefits to integrating CG into the RG process.\footnote{Link to our data and code will be provided on our project page \url{https://inklab.usc.edu/Reflect/}.}

\end{abstract}

\section{Introduction}\label{intro}
\begin{figure}[tb]
	\centering
    \vspace{-0.2cm}
	\includegraphics[width=0.9\columnwidth]{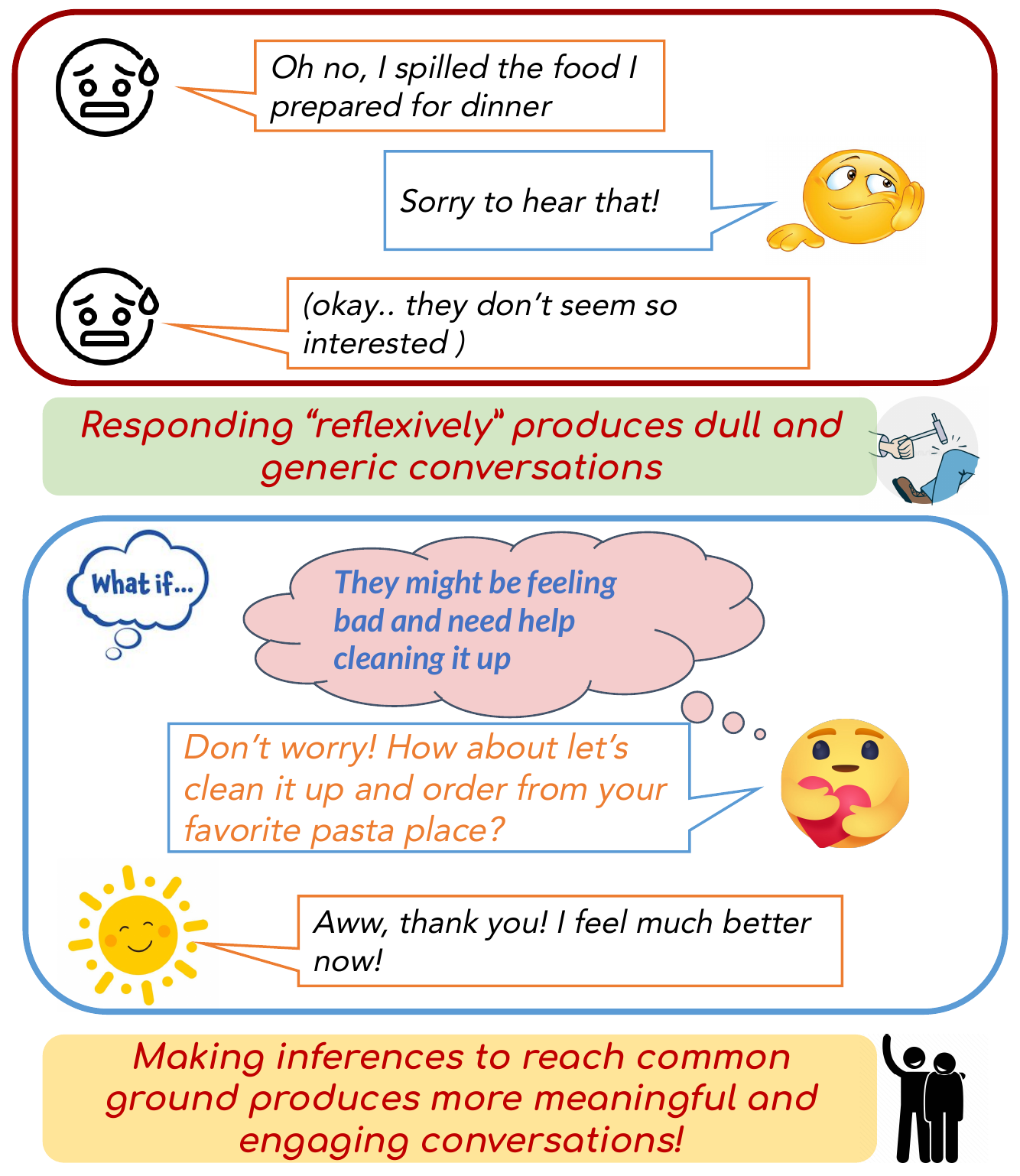}
% 	\vspace{-0.1cm}
	\caption{%\vspace{-0.5cm}
	{\textbf{A motivating example.} We aim to help RG models produce more \emph{human-like} responses instead of generic ones. We argue that integrating common ground by making inferences is crucial.}}
	  	\vspace{-0.5cm}
	\label{fig:motivation}
\end{figure}

% \textbf{[P1: Introduce common ground and current system limitations]}
Human communication is a collaborative effort~\cite{grice1975logic,allwood1976linguistic,bohm2004dialogue} where participants strive to achieve \textit{common ground} (CG), consisting of mutual beliefs and common knowledge~\cite{stalnaker1978assertion, clark1989contributing,clark1991grounding}. Conversational AI systems, while able to produce fluent texts, often generate generic and dull dialogue responses~\cite{serban2017hierarchical, zhao2017learning}, potentially because they do not explicitly model CG in communication (as illustrated in Figure~\ref{fig:motivation}). Specifically, existing models mostly follow a \emph{dialogue history $\rightarrow$ response} training paradigm since such data can be easily obtained in the wild, skipping an important middle step that \emph{builds common ground}, which naturally and universally exists in human communication, i.e., \emph{dialogue history $\rightarrow$ \textbf{common ground} $\rightarrow$ response}. Moreover, the same history can yield numerous responses, predicated on the CG and intent of the responder. We conjecture that the omission of modeling CG explicitly is a crucial bottleneck in RG models because they are directly trained to produce responses without learning \emph{how and why} those responses are uttered.
% \xiang{elaborate 2-3 more sentences about why NOT modeling common ground creates bottleneck to the current dialogue systems.}

\begin{figure}[tb]
	\centering
    \vspace{-0.3cm}
	\includegraphics[width=0.98\columnwidth]{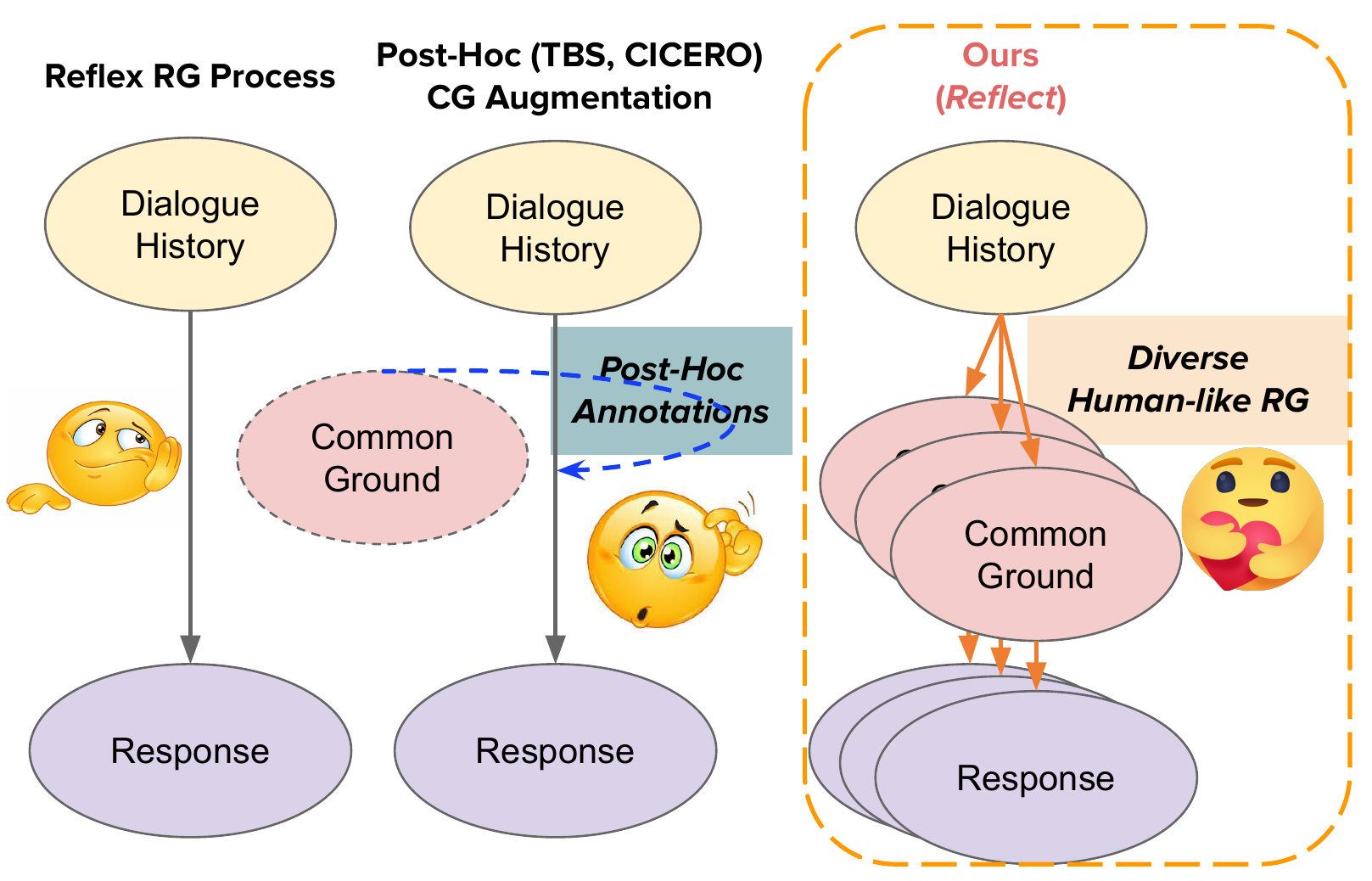}
% 	\vspace{-0.1cm}
	\caption{%\vspace{-0.5cm}
	{\small \textbf{Illustration of different RG approaches.} Common RG does not model CG explicitly, TBS~\cite{zhou2022TBS} and CICERO~\cite{ghosal2022cicero} post-hoc augments dialogues with CG, and we aim to follow natural human communication and first collect CG and then responses based on CG. We also factor in the diversity of plausible responses given a dialogue context that result from distinct CG.}}
	  	\vspace{-0.5cm}
	\label{fig:approach}
\end{figure}

% \textbf{[P2: Previous attempts on trying to model common ground and their limitations]}
Modeling common ground between speakers, however, is challenging due to its implicit and subjective nature during conversations\cite{clark1989contributing}.
Prior work on representing CG either mines noisy commonsense knowledge triples between dialogue history and existing responses~\cite{zhou2022TBS} or collects human inferences after reading the whole dialogue as a bystander~\cite{ghosal2022cicero}. Such approaches provide useful augmentation, but post-hoc analysis cannot mirror the generative process and intent of diverse human dialogue. 
Figure~\ref{fig:approach} illustrates three paradigms for RG.
%\xiang{Back this up with examples}
We argue that truly modeling this generative process requires (1) articulating CG prior to the response; (2) generating responses conditioned on CG; (3) differentiating response generation based on different types of CG. 

%We argue that they miss two crucial features of common ground: 1) common ground should be built \emph{before} making dialogue responses instead of matched to the dialogue in a post-hoc manner and \emph{by} the responder to help the RG process instead of a standalone reasoning step; 2) common ground is multi-dimensional in nature, meaning that there are various different ways to build it given a context, further leading to diverse sets of responses.

% \textbf{[P3: Present our approaches in response to limitations above (can be combined with P2 or P4]}
% In response to the limitations above and the integral role common ground plays in human communication, 

To this end, we formalize common ground in dialogues as \emph{inferences} made by \emph{one} participant to approximate potential beliefs shared by \emph{other} participants, as shown in Figure~\ref{fig:motivation}.
In this work, we instantiate inferences as \emph{question-answer (QA) pairs in natural language (NL)} such as \emph{``What might happen later?'' ``They might need to clean the floor''} to elicit others' beliefs, inspired by inquiry-based dialogic learning~\cite{bruner1961act,habermas1985theory,wells2000dialogic}.
Another critical aspect of CG is its multi-dimensional nature, i.e., given the same dialogue context, different plausible inferences can be made, which then lead to different responses.
% \xiang{before introducing the dataset, first articulate what is the desirable format/properties you propose to have, for studying/modeling common ground.}
Following these principles, we create a novel dialogue resource with multiple explicitly human-annotated common ground, each of which is further substantiated as a next-turn response continuing the conversations (an example of expanded CG and responses for one context shown in Figure~\ref{fig:turking}).

% For example, if your friend said: ``\emph{Oh no, I spilled the spaghetti I prepared for dinner!}'' you might first make the inference in your head that \emph{``What might happen later?'' ``They might need to clean the floor''} and then you might respond: ``\emph{Do you need a hand? I can help you find a mop}'', which is a much more engaging and specific response than what current RG systems might generate, e.g., ``\emph{I'm sorry to hear that.}''

% Pei: commented the following paragraph since I feel like directly going to data collection procedure reads more smooth to me, but @Jay let me know if you think we should put these back

% We formalize these intuitions about dialogue quality and common ground by performing a large-scale analysis of existing dialogue corpora and compare these with current state-of-the-art RG models and human-generated responses. We establish metrics to demonstrate when responses are specific to a context and respond to diverse common ground. Using these metrics, we characterize human-like dialogue responses and demonstrate that existing RG models fail to achieve human-like performance. Based on these findings, we collected a new dataset supporting more human-like RG.

% \textbf{[P4: Present our data collection and research questions for experiments]}
We design a two-stage data collection process by first asking crowdsourcing workers to answer different inference questions eliciting beliefs about CG (\emph{e.g., what is the speaker feeling right now?}) Answers rely on common sense, and adopt the point of view of the conversational respondent. We use these QA pairs to approximate various (non-exhaustive) inference dimensions to extend the common ground (e.g., empathy and event causality). Our second step converts these CG into dialogue responses by asking different workers to write a coherent response based on the answer/inference collected in the first stage. Our collected data \Reflect~contains 9k diverse responses from 600 dialogue contexts, based on 5 inference dimensions for CG.
% \xiang{Highlight some stats to give people an idea of scale and ``diversity of inference dims".}

% \pei{below needs changing}
% \xiang{Suggestion: Use two paragraphs, one for each part of the study. Each paragraph first describe the questions of interests and then summarize the findings and implications.}
Using~\Reflect, we first test our hypothesis that explicitly modeling CG and using CG to construct responses creates more engaging conversations.
We conduct human evaluation to compare the quality of responses between \Reflect~and ``\emph{reflex}'' style datasets and models in terms of sensibility, specificity, and interestingness. We find that, compared to reflex-prone human-written and machine-generated dialogues, our two-stage data collection process results in more responses that are sensible, specific, and interesting as rated by humans. This highlights limitations of existing data collection procedures and models trained on the data. 
% \xiang{We need to frame this argument more carefully. Some of the existing resources are collected from human-machine conversation, so we cannot say it's not realistic, but more about how the data should be collected to capture a more engaging and human-like conversations, and more diverse dialogue} \pei{changed}

Next, we look to study the potential of explicitly modeling CG in dialogue systems to help build models that can create more engaging conversations.
As a case study, we use the inference dimensions from \Reflect~ and test two simple ways to guide RG using CG.
%For example, we prompt GPT3 with the inference questions such as ``\emph{What might happen later?}'' in addition to the dialogue context. 
We surprisingly find that simple approaches such as appending an inference question to the dialogue context before the response in the few-shot (FS) in-context examples (from \Reflect) help GPT3-175B~\cite{brown2020language} generate almost 30\% more responses that are deemed sensible, specific, and interesting than vanilla FS learning GPT3 (no inference question). We demonstrate that, when prompted to ``\emph{think}'' about an inference question (approximated CG), large models such as GPT-3 can create more engaging conversations. We also find that such effect is only shown in large models like GPT-3 as we find BlenderBot-440M~\cite{roller2020recipes} benefits from fine-tuning on \Reflect, but appending inference questions does not further increase response quality.
% \xiang{What about findings on BlenderBot/models of moderate size? Good to share takeaways too.}

% we formulate two tasks to empirically analyze RG models' capabilities to \emph{build common ground} (make inferences) and \emph{use common ground} (generate inference-controlled responses) using our collected data. Furthermore, we propose novel automatic unsupervised evaluation metric to measure intrinsically the specificity of generated response candidates.

In summary, our contributions are as follows: 1) we operationalize theories about common ground and formalize them for dialogue; 2) we collect the first large-scale (9k responses) dialogue dataset with diverse responses guided by CG and release this resource to facilitate training and evaluation; 3) we show important limitations of existing dialogue data and RG models that detract from engaging communication; 4) we demonstrate that CG can dramatically improve RG quality even with simple prompting, boosting quality by 30\%. The resources and results from this work promise to enable the research community to create and evaluate common ground-aware RG models.

\section{Inference-Based Common Ground}\label{concept}
% \pei{Introducing the concept and intuition of inference-guided common ground and why this is important. Then introduce mathematical formulation.}

% \xiang{Good to add the illustrative figure about side-by-side comparison of the abstractive process for different RG procedure.}

We formally introduce the notion of \emph{common ground} in conversations as the implicit variable conditioned on dialogue history and provides conditions to the next-turn response. 
% Recent work such as~\citet{adolphs2021reason, zhou2022TBS, ghosal2022cicero,shuster2022language} have studied common sense or implicit and external knowledge in the context of RG and here we attempt to unify these related concepts and refer to them as \emph{common ground} and the action of ``\emph{thinking}'' in communication as \emph{grounding}, following psycholinguistic literature.

\subsection{Grounding in Communication} 
Successful collaborative communication activity relies on mutual understanding of shared knowledge and beliefs~\cite{clark1991grounding,bohm2004dialogue} called \textit{\textbf{common ground}}. However, due to \emph{least collaborative effort}~\cite{grice1975logic,clark1989contributing} where communication participants try to minimize the effort spent on contributing to the interaction, establishing CG relies on \emph{signals} other than the surface communication information (i.e., actual utterances in a conversation). While humans in face-to-face communication receive some information from \emph{non-verbal} signals such as gestures and facial expressions, virtual systems such as chatbots often do not have access to such signals. Thus, we argue that they have to rely heavily on another crucial way of getting signals for establishing CG: \emph{making inferences} based on the surface communication utterances and common sense, in order to approximate two humans talking to create engaging conversations.

Furthermore, building CG by making relevant inferences also connects closely with the ``\emph{dual process}'' theories of human reasoning~\cite{stanovich2000individual,evans2003two,kahneman2011thinking}. We argue that the ``\emph{reflexive}'' RG is mostly modeling ``\emph{System 1}'' which is intuitive and associative, but a more deliberative and logical ``\emph{System 2}'' is lacking.

\subsection{Formulating CG in Dialogue}\label{formulation}
Consider three high-level components in communication efforts: context $C$ (often materialized as dialogue history consisting of a sequence of $n$ contributions $C=c_1,...,c_n$), common ground $G$, and a new contribution continuing the context (often referred to as a ``\emph{response}'' $c_{n+1}$. Specifically, for common ground $G$, we focus on signals gained from inferences and thus materialize $G$ as a list of $m$ potential inferences $G=I_1,...,I_m$ conditioned on the context. We furthermore materialize each inference as a QA pair in NL $I_j=(Q_j, A_j)$ (examples included in Figure~\ref{fig:turking} between Stage 1 and 2). We use QA format to express inferences to mimic inquiry-based dialogic learning~\cite{bruner1961act,habermas1985theory,wells2000dialogic} and follow empirical evidence that neural models take in QA-format knowledge effectively~\cite{shwartz2020unsupervised, zhou2022TBS}.

\section{Collecting \emph{Reflect} Data}\label{data}
% \pei{Present the data construction pipeline, quality control, and analysis}

\begin{figure}[t]
     \vspace{-0.4cm}
    \hspace{-0.2cm}
	\includegraphics[width=1.1\linewidth]{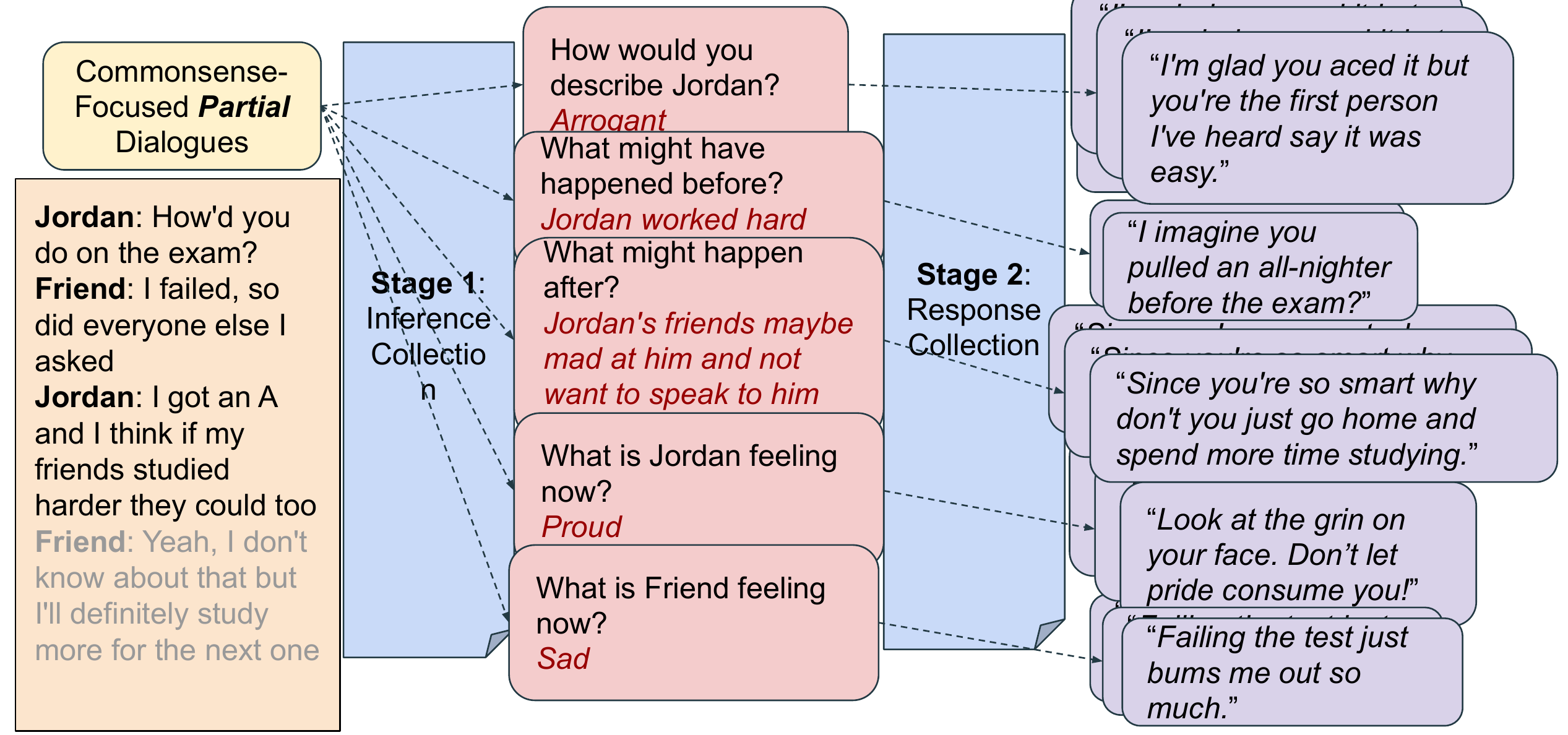}
 	\vspace{-0.4cm}
	\caption{%\vspace{-0.5cm}
	\small \textbf{Reflect collection procedure illustration.} We first collect CG materialized as inferences expressed in QA along different dimensions. Then for each QA pair, we collect multiple responses. 
% 	\yuchen{Use the blank space more wisely.}
% 	\xiang{compress this to a single column.}
	}
  	\vspace{-0.4cm}
	\label{fig:turking}
\end{figure}

% \xiang{Probably need to cut down to <1 page for this section. Move low-level details to appendix.}

Here we describe how we collect \emph{Reflect}, a novel large-scale dialogue dataset with diverse human-annotated inference-based CG and grounded responses. An overview of the procedure with examples are shown in Figure~\ref{fig:turking}. We first select base dialogues from a dataset that is constructed without considering CG and only has one plausible response for each context (\ref{pre_collection}). Then we aim to expand and collect multiple responses based on different inference dimensions.
We introduce a two-stage process to first crowdsource potential inferences people make in conversations (\ref{stage1}) and then ask a second cohort of workers to generate diverse responses based on the inferences (\ref{stage2}). 
We designed a two-stage data collection to 1) collect multiple, diverse responses based on each CG; 2) to allow response writers to validate CG as high quality, generic common sense inferences.
Finally, we include discussions of data quality assurance (\ref{qa}).
\subsection{Pre-Collection: Selecting Base Dialogue Turns for Expansion}\label{pre_collection}
Our first step is to select base dialogues and dialogue turns to expand on, in terms of both inference-based CG and more potential responses following the CG.
One important criterion for base turns is that they should not be ``\emph{social glue}'' turns such as ``\emph{You are welcome}'' in responding to ``\emph{Thank you!}''
We aim at expanding turns that have semantically-rich dialogue context, enabling different plausible inferences to be made.
After investigation of existing datasets, we use dialogues from Commonsense-Focused Dialogues~\cite{zhou-etal-2021-commonsense} that are converted to dialogues from SocialIQA~\cite{sap2019social} contexts. We chose this dialogue data because SocialIQA (crowdsourced from ATOMIC~\cite{sap2019atomic}, an if-then inferential commonsense knowledge base) contains everyday situations where people can make various inferences on. 
Then, to select what turns to expand on, we use simple heuristics and select the turn that has the largest semantic overlap with the event in SocialIQA using SentenceBERT~\cite{reimers-2019-sentence-bert}. 

\begin{table}[tb]
\centering
\resizebox{\columnwidth}{!}{
\begin{tabular}{c|l}
\hline
\textbf{Inference Dimensions} & \multicolumn{1}{c}{\textbf{Inference Questions}} \\ \hline
Attributes of Speaker         & How would you describe Speaker?                  \\
Potential prerequisites       & What might have happened before?                 \\
Potential consequences        & What might happen after?                         \\
Speaker Emotion States        & What is Speaker feeling now?                     \\
Responder Emotion States      & What is Responder feeling now?                   \\ \hline
\end{tabular}
}
\caption{ \small Inference dimensions and corresponding questions
% We can see that after our training, RG models can self-talk to generate \emph{proper} commonsense knowledge for around 75\% of the time. 
}
 \vspace{-0.3cm}
\label{tab:dimensions}
\end{table}

\subsection{Stage 1. Inference Collection}\label{stage1}
Our first goal is to collect potential inferences people might make (\emph{e.g}. ``\emph{they might be feeling bad}'') given conversation contexts $C$ to approximate \emph{common ground}. Each inference $I_j$ is further materialized as a QA pair $(Q_j, A_j)$ along multiple inference dimensions as formulated in Section~\ref{formulation}.

\paragraph{Inference Knowledge Schema}
We adopt inference dimensions from ATOMIC2020~\cite{Hwang2021COMETATOMIC2O} since it focuses on social commonsense inferences based on everyday scenarios. Specifically, we conduct a pilot study to choose 5 dimensions from the 14 dimensions, consolidating those that overlap (e.g., ``what might happen later'' and ``what would others likely want to do after'') in the context of dialogues. Our final five dimensions for conversation-based inference dimensions are shown in Table~\ref{tab:dimensions}.

\begin{figure}[tb]
\vspace{-0.4cm}
\centering
\begin{subfigure}[b]{0.4\textwidth}
\includegraphics[width=\textwidth,trim=0cm 0cm 0cm 0cm,clip=true]{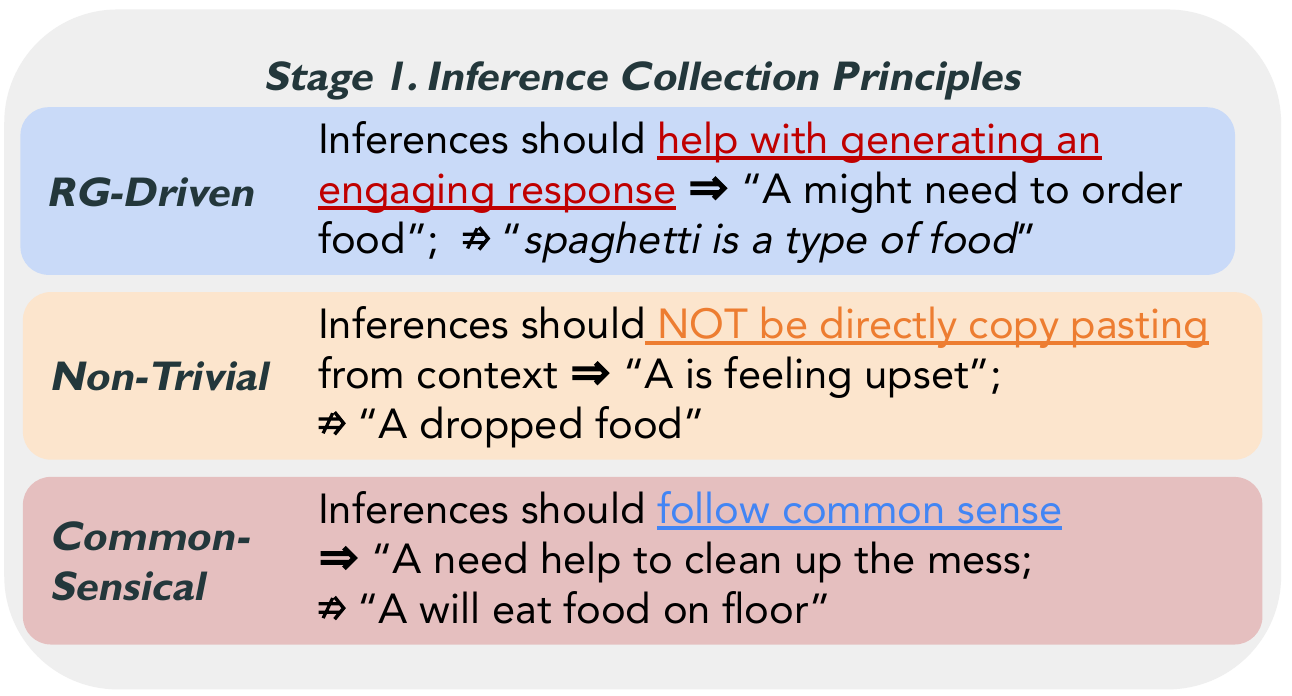}
\end{subfigure}
\begin{subfigure}[b]{0.4\textwidth}
\includegraphics[width=\textwidth,trim=0cm 0cm 0cm 0cm,clip=true]{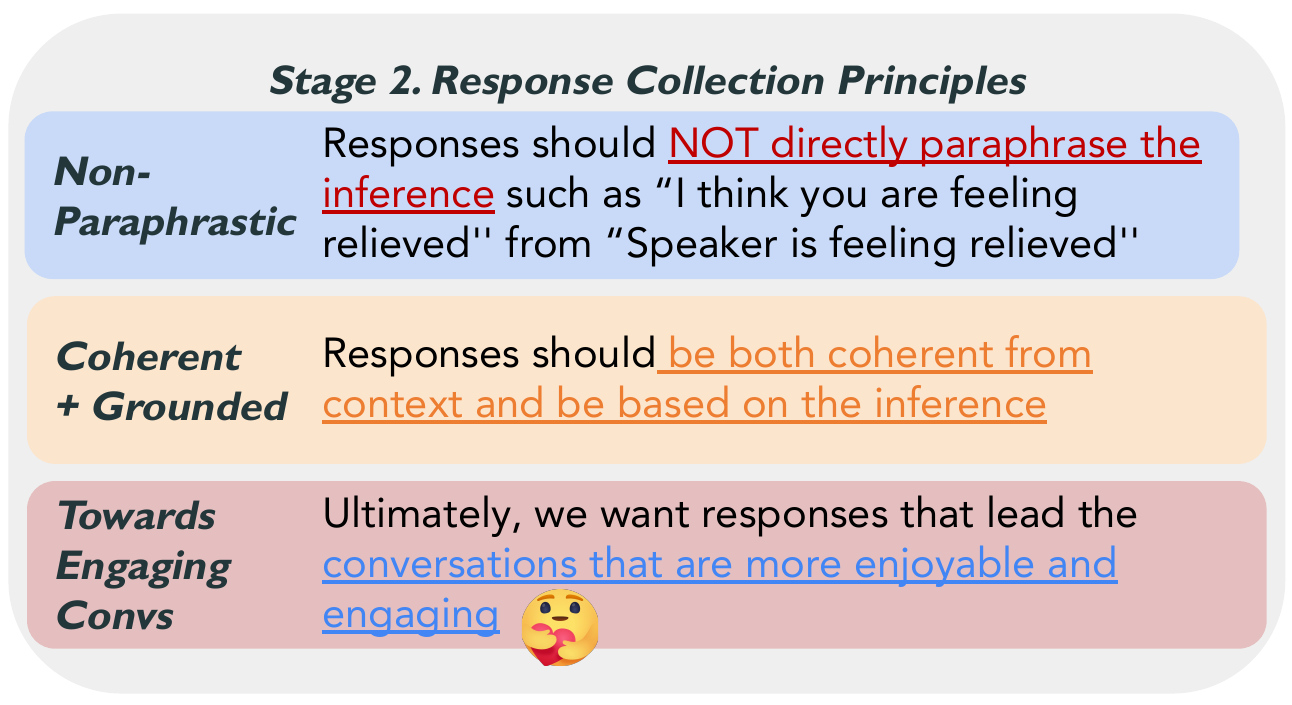}
\end{subfigure}
	\caption{%\vspace{-0.5cm}
	{\small Crowdsourcing principles for two-stage collection.}}
	  	\vspace{-0.3cm}
	\label{fig:stages}
\end{figure}

\paragraph{Crowdsourcing}
Our Stage 1 crowdsourcing task is: given a dialogue context, imagine that you are participating as the responder and write answers to the 5 inference questions (more details in Appendix).
We recruit a group of around 30 crowdsourcing workers from Amazon's Mechanical Turk platform (AMT) who are native English speakers and provide detailed feedback. 
Specifically, after carefully reading collected inferences from pilot studies, we provide feedback to turkers by stressing on several principles to make the inferences collected more closely approximate CG, shown in Figure~\ref{fig:stages}.

% \begin{figure}[tb]
% 	\centering
%     \vspace{-0.3cm}
% 	\includegraphics[width=0.98\columnwidth]{figures/stage2_principles.pdf}
% % 	\vspace{-0.1cm}
% 	\caption{%\vspace{-0.5cm}
% 	{\small Crowdsourcing principles for response collection }}
% 	  	\vspace{-0.5cm}
% 	\label{fig:stage2}
% \end{figure}

\subsection{Stage 2. Response Collection}\label{stage2}
After the first stage, we have collected 5 inferences (approximating CG) in the form of QA pairs for each dialogue context. Our next step is to collect next-turn responses given \emph{both} the dialogue context and the collected inference-based CG along different dimensions. To account for diversity in responses, for each dialogue context we ask three Turkers to write a next-turn response based on each of the given inferences, yielding 15 responses for each dialogue context. Similarly to Stage 1, we communicate our collection principles to workers to improve the collected data quality (Figure~\ref{fig:stages}). Both Stage 1 and Stage 2 UI and positive/negatives examples for workers are included in Appendix.

\begin{table}[tb]
\centering

\resizebox{\columnwidth}{!}{
\begin{tabular}{l|c|c|c}
\textbf{Resources} & \textbf{Source} & \textbf{Makes Sense} & \textbf{Relevant} \\ \hline
TBS~\cite{zhou2022TBS}                                 & ConceptNet                       & 83.7\%                                & 81.0\%                             \\
CICERO~\cite{ghosal2022cicero}                              & Human                            & 86\%                                  & \textbf{96\%}     \\
\Reflect~(Ours)                            & Human                            & \textbf{93\%}        & \textbf{96\%}    
\end{tabular}
}
\caption{ \small Human evaluation on \textbf{inference (CG) quality}. We compare inferences from three resources and compare their sensibility and relevance to dialogue context.
% We can see that after our training, RG models can self-talk to generate \emph{proper} commonsense knowledge for around 75\% of the time. 
}
 \vspace{-0.5cm}
\label{tab:data_compare}
\end{table}

\subsection{Quality Control and Analysis}\label{qa}
\paragraph{Quality check for Inference Collection}
In our second stage for response collection, we ask workers an additional question before writing a response: ``\emph{do you think the shown inference answer is a valid reaction from the responder?}'' as a way to check the quality of the first collection stage results. We find that less than 7\% (200/3000) of the inferences are deemed implausible by second stage workers and only keep the inferences where most workers agree that the inferences are plausible. 
\paragraph{Quality check for Response Collection}
To check quality for our stage 2 response results, we randomly sampled around 5\% of collected responses (500) and conduct a manual in-house check for two criteria: 1) is it a sensible continuation from the dialogue context? and 2) is the response based on the inference given? We find that around 93\% of the responses are a sensible continuation and 89\% are following the inferences given.
Further human ratings of our collected grounded dialogue responses showing that our data improves the sensibility, specificity, and interestingness aspects compared to the base responses are included and discussed in Section~\ref{reflex_limitations}.
\paragraph{Comparison to prior work on representing CG}
We compare CG inferences from \Reflect~with TBS~\cite{zhou2022TBS} and CICERO~\cite{ghosal2022cicero}, two prior work that aims to represent CG in dialogues using either ConceptNet~\cite{speer2017conceptnet} knowledge triples or post-hoc human annotations, respectively. Note we only compare inferences (CG) since neither collects new dialogue responses grounded in the inferences, and only consider a single response per context.
Comparison results on sampled 100 inferences for each resource are shown in Table~\ref{tab:data_compare} where we find that inferences in \Reflect~are rated as make more sense and relevant to dialogue context than the prior dataset.

% \subsection{Analysis}

\section{Limitations of Reflex-Prone Dialogue Data and Models} \label{reflex_limitations}

Most existing open-domain dialogue datasets are either crowdsourced by workers who do not have strong incentives to create engaging conversations~\cite{rashkin2019towards,zhou-etal-2021-commonsense} or crawled from language learning websites and exams~\cite{li2017dailydialog,cui2020mutual}. Both lack explicit CG.
These collection processes can fail to capture engaging human-like conversations through under-specified response criteria. 
Accordingly, RG models trained on these data may mimic generic patterns. 
This section aims to demonstrate such limitations by comparing responses from \Reflect~with responses from both the original dialogue dataset we expand on and models trained on the \emph{dialogue history $\rightarrow$ response} regime. 

\begin{table}[tb]
\centering

\resizebox{\columnwidth}{!}{
\begin{tabular}{c|c|c}
\hline
\textbf{Dimensions}                         & \textbf{Positive Examples}                                                                                                  & \textbf{Negative Examples} \\ \hline
\textit{Sensibleness}                       & That's too bad!                                                                                                             & Thank you.                 \\ \hline
\textit{Specificity}                        & Did you spill it in the kitchen? Let me help!                                                                               & Do you need help?          \\ \hline
\textit{Interestingness}                    & \begin{tabular}[c]{@{}c@{}}It's actually blessing in disguise, \\ wanna guess why?\end{tabular}                             & Let's eat something else.  \\ \hline
\multicolumn{1}{l|}{\textit{Quality (SSI)}} & \begin{tabular}[c]{@{}c@{}}It's blessing in disguise, since I ordered \\ extra from your favorite pasta place!\end{tabular} & All above                  \\ \hline
\end{tabular}
}
\caption{ \small Evaluation dimensions for RG with examples (dialogue context from Figure~\ref{fig:motivation}).
% We can see that after our training, RG models can self-talk to generate \emph{proper} commonsense knowledge for around 75\% of the time. 
}
 \vspace{-0.5cm}
\label{tab:SSI}
\end{table}

\subsection{Human Evaluation Dimensions-SSI}\label{human_eval}
% Our DA focuses on statistical attributes of generated response sets and adopt intuitions of what make human responses engaging. However, we still need 
We evaluate the \emph{quality} of each response by head-to-head comparing across systems along several evaluation criteria.
We follow the protocol used by LaMDA~\cite{thoppilan2022lamda} and measure SSI: sensibleness, specificity, and interestingness.
Examples of positive and negative responses are shown in Table~\ref{tab:SSI}. 
Our assumption is that responses that contribute to more engaging conversations should satisfy \emph{all} three dimensions and we refer to them as \emph{\textbf{quality} responses}. 
We do not consider automatic metrics since they do not yet reliably replace human judgements on open-ended responses, especially for fine-grained evaluation dimensions.

% The average fleiss-kappa agreement for sensible, specific, and interesting scores range from 0.2 to 0.4 with specificity the highest, indicating fair agreement according to one interpretation~\cite{fleiss1971measuring}. The subjectivity nature of evaluating response and prior work of similar agreement~\cite{} indicate open challenge to evaluate dialogue responses.

\subsection{Comparing Original vs Reflect Responses}
First, we compare the quality of responses in previous dialogue datasets with our \Reflect~responses to analyze the effects of explicitly incorporating CG in \emph{human RG}.
Here we present results by adopting the aforementioned evaluation protocol on human dialogues, both from the original base dialogues~\cite{zhou-etal-2021-commonsense} and from our Reflect dataset, derived from the same dialogues. 
% We use the best-performing sentence embedding model (mp-net-v2)~\cite{reimers-2019-sentence-bert} that converts each sentence to a 748-dimension vector. 
We sampled 300 dialogue contexts and asked 3 crowdsourcing workers to rate the three SSI criteria, using majority voting to get final scores (Fleiss-kappa~\cite{fleiss1971measuring} agreement is around 0.67). We compare the original next-turn response from the contexts with a randomly sampled one from our Reflect responses. 

\begin{figure}[tb]
    % \vspace{-0.2cm}
% 	\hspace{-0.5cm}
	\includegraphics[width=\columnwidth]{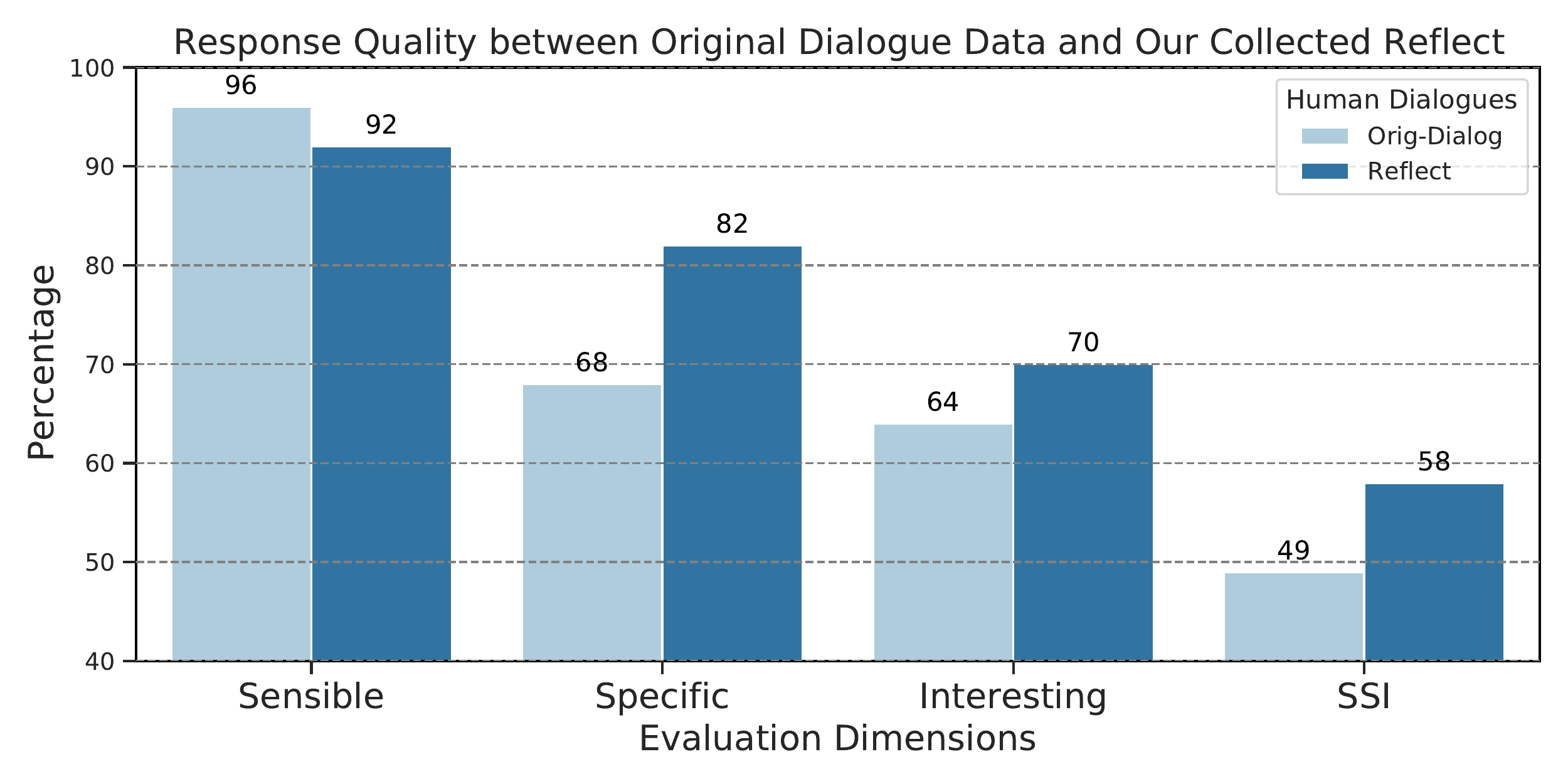}
% 	\vspace{-0.7cm}
	\caption{%\vspace{-0.5cm}
	\small \textbf{Human evaluation} comparing human dialogues: original data and our collected Reflect.	%We find replacing generated knowledge to random noisy knowledge hurts grammar, coherence, engagingness, and common sense aspects significantly.  
	%\xiang{1) Update the figure per our discussion; 2) can be made to double column; flatten the figure (less height) and make each bar thicker}.
	}
 	\vspace{-0.3cm}
	\label{fig:Reflect_VS_Orig}
\end{figure}

\paragraph{\Reflect~contains more specific and interesting responses than original dialogues}
From human evaluation shown in Figure~\ref{fig:Reflect_VS_Orig}, we observe that our collected \Reflect~data, consists of dialogue responses that are on average more specific (20\%) and interesting (13\%) than the original data, while having slightly lower sensibility (4\%) ratings. One possible contributor to the lower sensibility may be 2-stage collection where a new worker continues dialogues constrained by a specific inference generated by another person.
Specifically, when comparing the percentages of responses that satisfy all three criteria, \emph{i.e.,} quality responses, we find that there are substantially more (18\%) such responses in \Reflect~than in original data.
This observation raises an interesting question: ``\emph{do existing dialogue training datasets capture high quality dialogues?}'' 
Without sensible, specific, and interesting responses to learn from, RG models will necessarily be limited in the quality of their output.
% Admittedly, humans do not always engage in meaningful conversations as we can see in the post-hoc human ratings here. 
% However, we argue that we should improve our data collection procedure, such as involving common ground explicitly, to capture more engaging human communication, so that we can steer RG models towards more desirable communicators.
% We also argue that these features make our data more desirable to train and evaluate RG models since we want models to produce more meaningful and engaging conversations, which are covered in these two dimensions.

\subsection{Comparing Reflex RG vs Reflect Data}
We now compare \Reflect~with RG models trained on dialogue data that lacks explicit CG and to directly generate an utterance given a context.
\paragraph{Reflexive model baselines}
Specifically, we consider models from two categories: medium-sized RG models pre-trained on dialogue data such as BlenderBot (440M parameters)~\footnote{\url{https://parl.ai/projects/recipes/}}~\cite{roller2020recipes} and large-sized language models (LLM) pre-trained on general texts such as GPT3-DaVinci (175B parameters)~\footnote{\url{https://beta.openai.com/docs/models/gpt-3}}~\cite{brown2020language}. 
We directly use off-the-shelf Blender since it is pre-trained on dialogue data (\textbf{Blender}).
For GPT3-175B, we apply \emph{few-shot in-context} learning by providing 3 examples of dialogue context and response from existing data (\textbf{GPT3-FS}). We manually examine these responses to ensure their quality as demonstrating examples. Then we present a dialogue context from our test data and prompt GPT3 to generate a next-turn response. More details in Appendix~\ref{appendix_data}. 

\begin{figure}[tb]
    % \vspace{-0.2cm}
 	\hspace{-0.5cm}
	\includegraphics[width=\columnwidth]{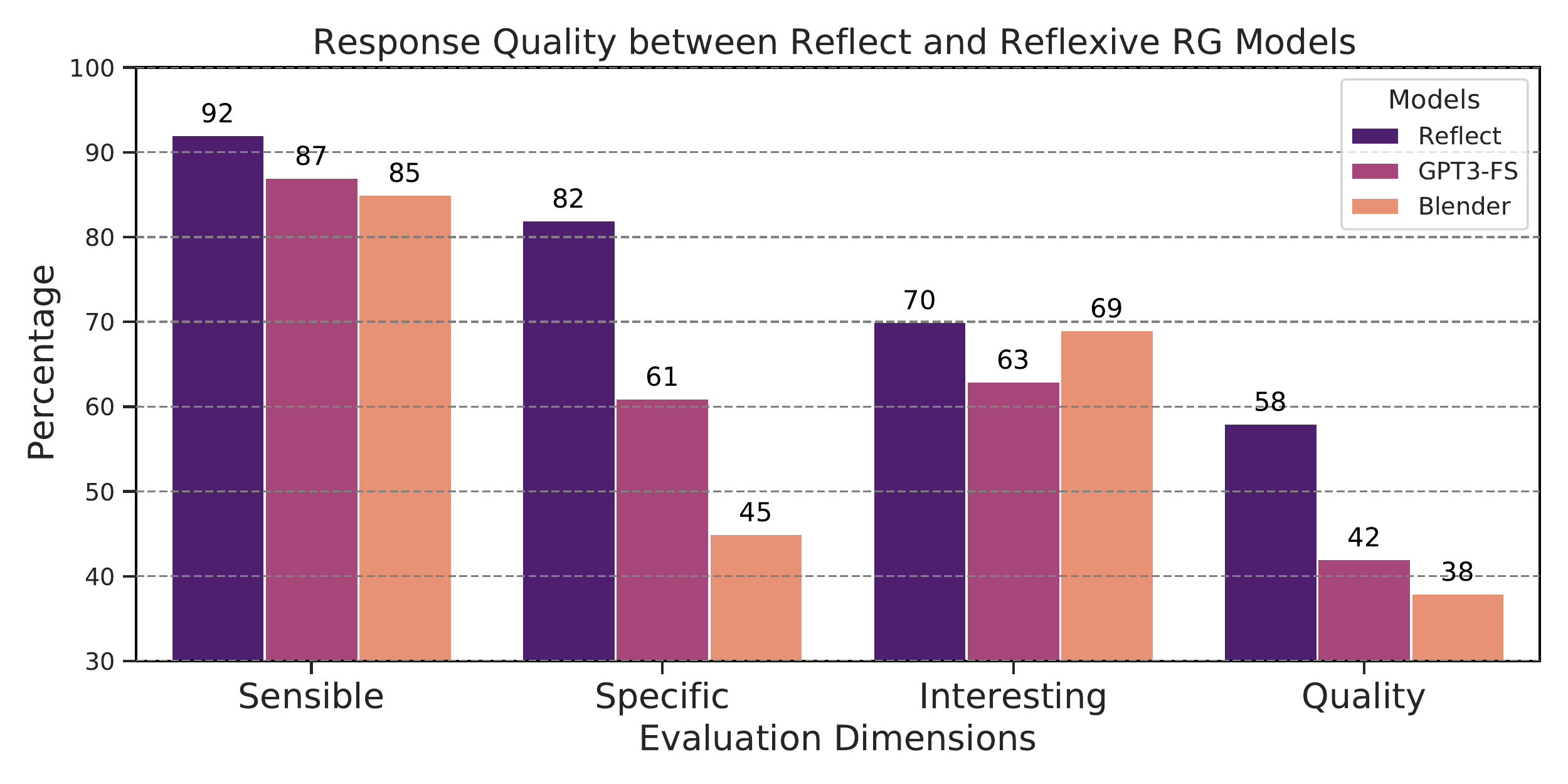}
% 	\vspace{-0.7cm}
	\caption{%\vspace{-0.5cm}
	\small \textbf{Averaged performance comparing \Reflect~with reflexive models}. Predictably, RG models trained to directly output responses from dialogue context using no-common ground responses have a large gap in response quality compared to human responses guided by inferences.
	}
 	\vspace{-0.3cm}
	\label{fig:Reflect_VS_RG}
\end{figure}

\paragraph{Models with no common ground struggle}
Unsurprisingly, as shown in Figure~\ref{fig:Reflect_VS_RG}, we find a similar trend as comparing \Reflect~with original dialogue data: both BlenderBot-FT and GPT3-FS generate much fewer \emph{quality} responses (53\% and 38\%, respectively) that satisfy all criteria and particularly on specificity. This further supports the hypothesis that RG models that learn from no-grounding dialogue responses struggle to capture what constituted meaningful conversations.


\section{A Little CG Goes a Long Way} \label{CG_benefits}
\begin{table*}[tb]
\centering
\resizebox{\linewidth}{!}{
\begin{tabular}{|c|ll|}
\hline
\textbf{Dialogue History}             & \multicolumn{2}{l|}{\begin{tabular}[c]{@{}l@{}}A: One of my patients keeps forgetting to take his medicine on time.\\ B: That's not good, how can you help someone like that? Why did you have to quit?\\ A: I've been calling him everyday to make sure he takes his medicine, I have noticed his health is improving too.\end{tabular}}                                                                                                     \\ \hline
\textbf{Orig Response}                & \multicolumn{2}{l|}{B: You're a great nurse!}                                                                                                                                                                                                                                                                                                                                                                                                 \\ \hline
\multirow{5}{*}{\textbf{Reflect}}     & \multicolumn{1}{l|}{\begin{tabular}[c]{@{}l@{}}Q: How would you describe A? \\ Ans: A is kind, patient and persistent.\end{tabular}}                                                                       & \begin{tabular}[c]{@{}l@{}}B: You are really determined and wonderful with your patients. Not many doctors would \\ take the time every day to do something like that.\end{tabular}                                              \\ \cline{2-3} 
                                      & \multicolumn{1}{l|}{\begin{tabular}[c]{@{}l@{}}Q: What might have happened BEFORE?\\ Ans: A noticed a patient's health wasn't improving with\\ their medicine.\end{tabular}}                               & \begin{tabular}[c]{@{}l@{}}B: What made you notice he wasn't taking it? Did they need a refill by your notes but still \\ have a lot left? Or did you notice they weren't better at their last appointment?\end{tabular}         \\ \cline{2-3} 
                                      & \multicolumn{1}{l|}{\begin{tabular}[c]{@{}l@{}}Q: What do you think might happen AFTER?\\ Ans: A will contact a family member of the patient to help\\ their patient remember to take meds..\end{tabular}} & \begin{tabular}[c]{@{}l@{}}B: How will you handle it going forward? Can you keep up with that, or will you see if \\ someone else close to them can help?\end{tabular}                                                           \\ \cline{2-3} 
                                      & \multicolumn{1}{l|}{\begin{tabular}[c]{@{}l@{}}Q: What do you think A is feeling?\\ Ans: Satisfied, successful and determined.\end{tabular}}                                                               & \begin{tabular}[c]{@{}l@{}}B: You must be feeling really confident and strong right now. You're a doctor who goes \\ above and beyond for your patients. I've never met another who would do this.\end{tabular}                  \\ \cline{2-3} 
                                      & \multicolumn{1}{l|}{\begin{tabular}[c]{@{}l@{}}Q: What do you think B is feeling?\\ Ans: Impressed and optimistic for A.\end{tabular}}                                                                     & \begin{tabular}[c]{@{}l@{}}B: Wow, that's so amazing. I can't believe you've done at all that. Most doctors would just \\ assume it's the patients responsibility. I am confident you'll get him better after this.\end{tabular} \\ \hline
\textbf{GPT3-FS-Top5}                 & \multicolumn{2}{l|}{\begin{tabular}[c]{@{}l@{}}B: That is amazing. You have to be a great nurse to help people like that.\\ B: You are a true caregiver and I'm glad you're taking care of your patients so well.\\ B: That's good, once he stops forgetting you can give him back his phone and he can do it on his own.\\ B: Well, you are a good doctor.\\ B: That is good.\end{tabular}}                                                  \\ \hline
\multirow{5}{*}{\textbf{GPT-FS-InfQ}} & \multicolumn{1}{l|}{Q: How would you describe A?}                                                                                                                                                          & B: You are amazing! I could never do that job.                                                                                                                                                                                   \\
                                      & \multicolumn{1}{l|}{Q: What might have happened BEFORE?}                                                                                                                                                   & B: That's really good that you took the initiative to make sure he was taking his medicine                                                                                                                                       \\
                                      & \multicolumn{1}{l|}{Q: What do you think might happen AFTER?}                                                                                                                                              & B: You are really dedicated to your job. I hope you get a raise soon.                                                                                                                                                            \\
                                      & \multicolumn{1}{l|}{Q: What do you think A is feeling?}                                                                                                                                                    & B: It's great that you are helping him out and that his health is improving.                                                                                                                                                     \\
                                      & \multicolumn{1}{l|}{Q: What do you think B is feeling?}                                                                                                                                                    & B: I'm glad that the patient is improving, but it's a lot of work for you.                                                                                                                                                       \\ \hline
\end{tabular}
}
\caption{
\small \textbf{Randomly sampled dialogue example to qualitatively compare responses from different resources and models.} Compared to original and GPT3 top-5 responses which mostly express ``\emph{You are a great nurse/doctor}'', \Reflect~responses based on the inference QA pairs are much more specific and engaging. We find that GPT3-InfQ, though only guided by an inference question, generates less genric responses such as ``\emph{... I hope you get a raise soon.}''
% \xiang{@Pei I didn’t get enough context on how this exp was done and why we select the two baselines here. The naming of the method also confuses me --- I don’t know what exactly our method is based on the current information given.} Pei: changed naming and added explanation in writing
%\xiang{1) only show complete corruption; 2) the name ``End-to-end" sounds vague: what about just ``DialogGPT" or ``finetuned DialogGPT"}
}
\label{tab:more_examples}

\end{table*}

\begin{figure}[tb]
    % \vspace{-0.2cm}
% 	\hspace{-0.5cm}
	\includegraphics[width=\columnwidth]{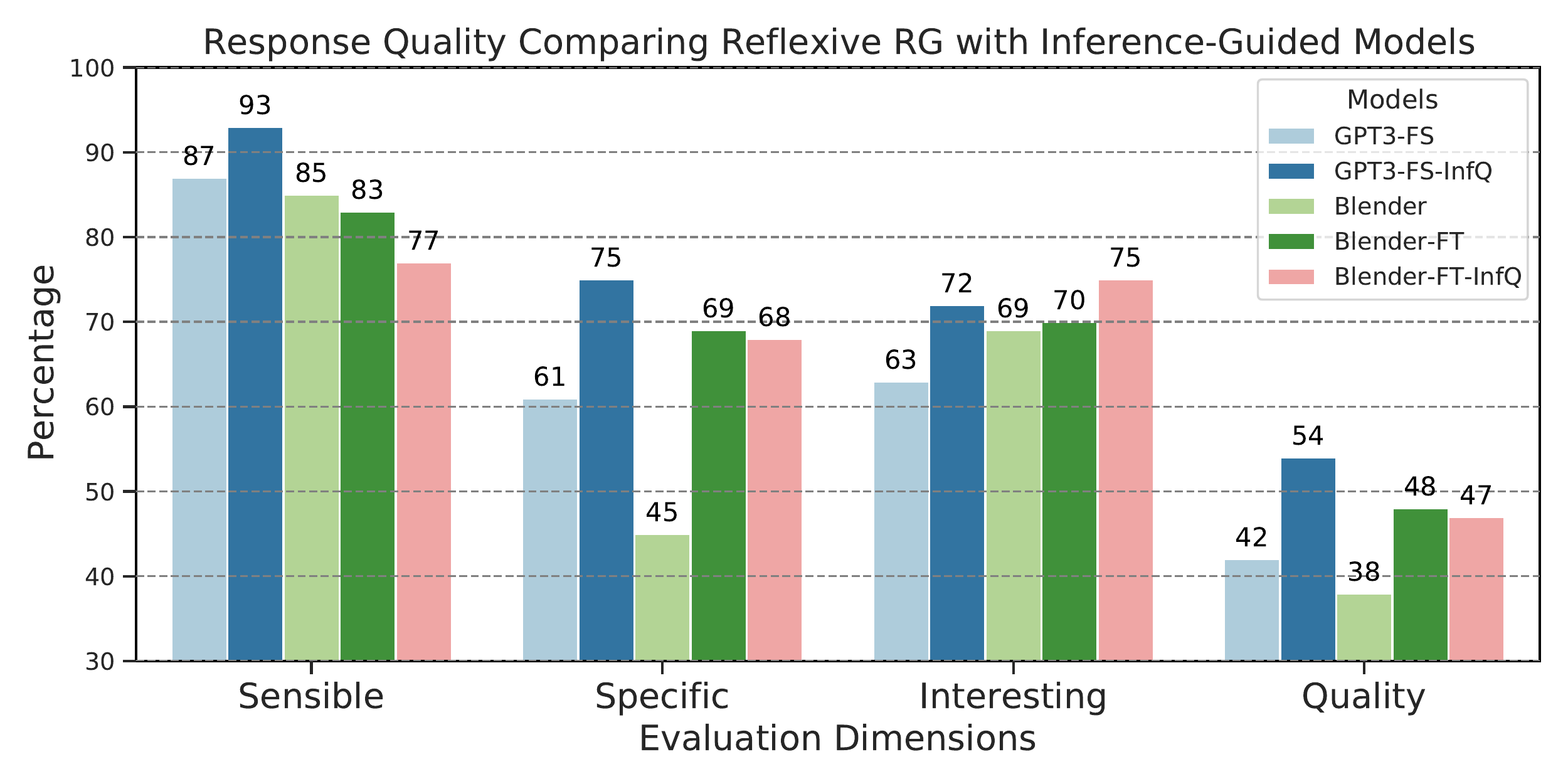}
% 	\vspace{-0.7cm}
	\caption{%\vspace{-0.5cm}
	\small \textbf{Averaged performance comparing before and after reflexive models leveraged inference-guided \Reflect~data}. We find that inference-based common ground prompting helps GPT3-175B significantly, even comparable to human responses from Figure~\ref{fig:Reflect_VS_Orig}. Blender also gained much improvement from pre-trained after fine-tuning on \Reflect, however we find no significant effects on appending inference questions.
	}
 	\vspace{-0.3cm}
	\label{fig:effects_CG}
\end{figure}

% Here we investigate and present results around three central questions: 1) do off-the-shelf dialogue systems understand the multiple inference dimensions for building common ground during communication? 2) given human-written inferences, can models generate or choose proper responses that are valid and based on the inference? 3) can models successfully mimic human communication process and generate more specific and interesting responses by first building common ground and then responding?
After showing that explicitly integrating inference-based CG helps \emph{humans} produce more specific and interesting dialogue responses, we now test if this also holds for neural RG models. 
We take the non-exhaustive inference dimensions we used in \Reflect~as case studies to see how CG could improve the quality of existing RG systems' responses, in terms of the SSI human evaluation~\cite{thoppilan2022lamda}.

\subsection{Experiment Setup}
\paragraph{Inference-Guided reflect models}
We attempt to shift models from ``\emph{reflexive}'' RG to ``\emph{reflective}'' RG by taking into account of plausible inferences that humans use to build common ground during communication.
Since both BlenderBot and GPT3 are trained to generate responses directly without integrating common ground, a non-trivial challenge is how to adapt them to use inference-based common ground before RG. 
Here we present our two intuitive and simple approaches.

For BlenderBot-440M, we follow the common practice of fine-tuning models to adapt to a new task format. We split our Reflect data into 60/10/30 for train/valid/test and first fine-tune BlenderBot-440M (\textbf{Blender-FT}) on only the collected responses to show potential benefits of training from inference-guided human responses.
Then we fine-tune BlenderBot but modify the training task from outputting responses from contexts to inference-guided RG.
Inspired by modular generation in dialogue RG~\cite{adolphs2021reason,zhou2022TBS,shuster2022language}, our training task is: given dialogue context and one of the five inference dimension questions, generate the answer as well as the response collected in Reflect (\textbf{Blender-FT-InfQ}, indicating that the model is given the \emph{Inference Question}). More details in Appendix~\ref{appendix_implementation}.

For GPT-175B, we follow the few-shot in-context learning approach with one small addition in input: we append the dialogue context with an inference question and ask the model to generate a response. Our pilot studies show that GPT3 tends to generate directly an answer to the question, not a next-turn response to the dialogue context, thus we format the question into a prompt for GPT3 and stress that the end goal is RG. Specifically, we append the text ``\emph{Think about this when responding: }'' and then one of our inference questions after the dialogue context to prompt GPT3 to generate a response by reflecting on the questions (\textbf{GPT3-FS-InfQ}). Illustrative figures for prompting GPT3 are shown in Appendix~\ref{appendix_data} Figures~\ref{fig:GPT3_infq}.

To compare and analyze the effects of each inference dimension, we randomly sample one response for each of the five inference dimensions for GPT3-FS-InfQ and Blender-FT-InfQ and take their average. For GPT3-FS, Blender, and Blender-FT, we pick the \emph{top 5} responses generated using their default decoding strategy (beam search for GPT3 and nucleus sampling for Blender) and aggregate their evaluation results. In total, we evaluate 250 responses from \emph{each} model following the procedure in Section~\ref{human_eval}.

\subsection{Experimental Results}
% \paragraph{Reflexive GPT3 responses lack specificity and interestingness, behind BlenderBot}
% When comparing the powerful and massive GPT3-DaVinci model with human written dialogues, we find that previously argued key limitations of neural RG models still hold: they can produce fluent but dull and boring responses. Specifically, GPT3, trained in a reflexive way to do RG (both in pre-training data and our in-context examples), generates much less specific and interesting responses compared with Reflect, but impressively performs on par with the original dialogues.
% On the other hand, BlenderBot fine-tuned on Reflect responses impressively generates good quality responses.

\paragraph{Prompting GPT3 to ``\emph{think}'' about common ground improves response quality by 30\%}
Figure~\ref{fig:effects_CG} presents results when comparing models that has no access to inference-guided \Reflect~data with those that do.
We test the hypothesis that whether guiding RG models with inference questions about common ground is helpful for generating more human-like responses. We find that with inferences, GPT3-FS-InfQ outperforms GPT3-FS on \emph{all} evaluation dimensions. Specifically, inference-guided GPT3 produces almost 25\% more specific and 30\% more quality responses. Moreover, 54\% quality (sensible, specific, and interesting) responses already surpasses quality of human-written responses in original dialogues (49\%), but still lags behind \Reflect~(58\%) as shown in Figure~\ref{fig:Reflect_VS_Orig}.

\paragraph{Fine-tuning Blender on \Reflect~generates 26\% more quality responses}
For BlenderBot-400M, we find that fine-tuning on inference-guided human responses from \Reflect~helps generate almost 50\% more specific and 26\% more quality responses.
In contrast to GPT3, BlenderBot with inference-guided fine-tuning does not seem to improve much.
We speculate that model size might play a role in how much model is influenced by CG inferences, leaving future work for more  inference-customized fine-tuning on moderate-sized models.
\begin{figure}[tb]
\vspace{-0.4cm}
\centering
\begin{subfigure}[b]{0.5\textwidth}
\includegraphics[width=\textwidth,trim=0cm 0cm 0cm 0cm,clip=true]{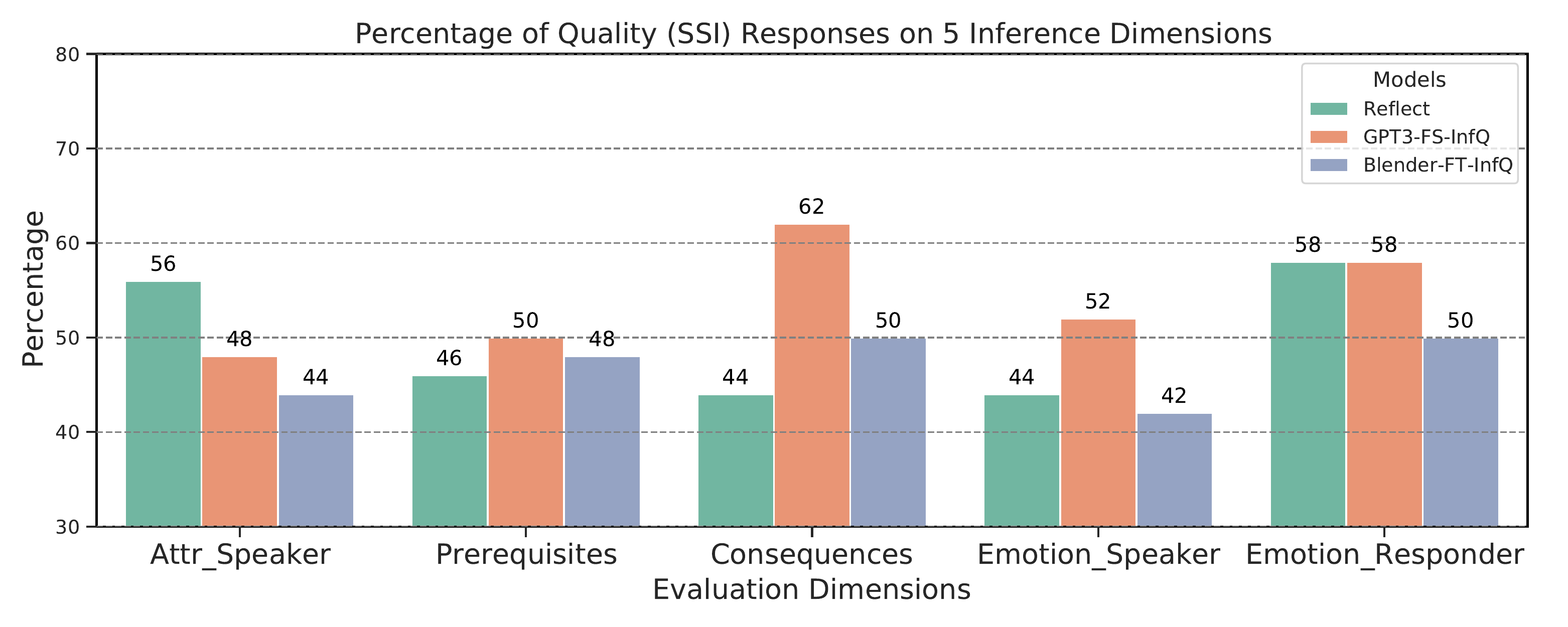}
\end{subfigure}
% \begin{subfigure}[b]{0.4\textwidth}
% \includegraphics[width=\textwidth,trim=0cm 0cm 0cm 0cm,clip=true]{figures/RQ_InfDim2.pdf}
% \end{subfigure}
% \begin{subfigure}[b]{0.4\textwidth}
% \includegraphics[width=\textwidth]{figures/RQ_InfDim3.pdf}
% \end{subfigure}
% \begin{subfigure}[b]{0.4\textwidth}
% \includegraphics[width=\textwidth]{figures/RQ_InfDim4.pdf}
% \end{subfigure}
% \begin{subfigure}[b]{0.4\textwidth}
% \includegraphics[width=\textwidth]{figures/RQ_InfDim5.pdf}
% \end{subfigure}
% \vspace{-0.2cm}
\caption{\small \textbf{Response evaluation separated by inference dimensions}. We find that GPT3-FS-InfQ generate better responses than humans on the potential consequences dimension while generates worse on attributes.
% \xiang{Put all figures into single column space.}
}

\label{fig:inf_separated}
\vspace{-0.4cm}
\end{figure}

\subsection{Analysis}

\paragraph{Which inference dimension helps models the most (and which the least)?}
Figure~\ref{fig:inf_separated} shows the percentages of quality responses separated by the inference dimension we use to prompt humans and models. 
Interestingly, we find that on some dimensions, GPT3-FS-InfQ can produce significantly better responses than human responses from \Reflect, especially event-based: ``\emph{What might have happened before}'' and ``\emph{what might happen after?}'' and emotion-based CG about the other speaker ``\emph{What is A (speaker1) feeling now?}''.
However, on ``\emph{How would you describe A}'', humans responses grounded on this question are much better. This dimension-specific analysis provides evidence that neural models' capability to generate quality responses may depend on what \emph{types of CG} we use to guide them.

\paragraph{Prompting GPT3-175B with complete human inferences}
To show how well GPT3 can make use of complete human-annotated common ground, we further append the inference answer after the question from \Reflect~data and prompt GPT3 to generate a response given the fully materialized common ground. As expected, we observe further improvements in response quality especially in specificity (15\% more) and general quality (16.7\% more). 
This analysis shows promises to make reflect-style models produce better responses by providing quality inference answers for CG.

% Specifically, TBS uses the same base dialogues~\cite{zhou-etal-2021-commonsense} where we compare in Figure~\ref{fig:Reflect_VS_Orig} and CICERO uses DailyDialog~\cite{li2017dailydialog}, MuTual~\cite{cui2020mutual}, and DREAM~\cite{sun2019dream}, which all

% \section{Analysis} \label{analysis}
% \input{sections/6_analysis}

\section{Related Work}\label{rel_work}

%%%

% \xiang{Related work needs some rewriting: the main lines of related work:
% 1) dialogue generation and especially knowledge-augmented RG;
% 2) other related study on method for knowledge-augmented NLU;
% 3) very briefly on implicit knowledge generation efforts.} updated

We have presented discussion of previous work representing CG~\cite{ghosal2022cicero,zhou2022TBS} in Section~\ref{intro} and relevant communication theory and psycholinguistic literature in Section~\ref{concept}. Here we provide additional discussions.
Recent advances on neural RG models mainly focused on fine-tuning large pre-trained transformer models~\cite{zhang-etal-2020-dialogpt,roller2020recipes,thoppilan2022lamda} on huge number of dialogue data. However, few of the data provides explicit common grounding.
Modular RG~\cite{adolphs2021reason, shuster2022language} aims to generate relevant knowledge first by retrieving from the web and the generate knowledge-grounded responses. Compared to these work, we focus on inferences based on common sense instead of external knowledge. 
Another closely related work by \citet{cho-may-2020-grounding} examined incorporating dialogue data with techniques from improvisational theater to teach models to implicitly build common ground. 

\section{Conclusion}\label{conclusion}
We introduce \Reflect, a dataset with diverse inference-grounded responses inspired by CG and communication theories. We carefully design our two-stage collection process and apply quality control. Then we demonstrate limitations of existing dialogue data and models trained on it. Finally, we present promising signs that guiding models with CG results in more engaging conversations. We hope to encourage more work on improving RG quality by looking at how humans use CG and adapt the communication process to machine learning models. Future directions include providing a \emph{ranking} of inference dimensions depending on dialogue context and train models to generate responses following the most suitable dimension. \Reflect~also enables potential automated metrics to evaluate response since more responses per dialogue might help gauge the plausible response space given a context.

\section*{Acknowledgments}
We thank anonymous reviewers for providing insightful feedback along with Brendan Kennedy, Peifeng Wang, and members from INK and JAUNTS lab. This research is supported in part by the DARPA MCS program under Contract No. N660011924033, the Defense Advanced Research Projects Agency with award NSF IIS 2048211, NSF SMA 182926, and support from Google.

\section*{Ethics and Broader Impact}
We collect a new dialogue dataset in English, which benefits English speakers more. We use Amazon Mechanical Turk to recruit crowdsourcing workers and we pay workers over \$15/hour on average, well above the highest state minimum wage and engage in constructive discussions if they have concerns about the process. We also give each annotation instance enough time so that we do not pressure annotators.
In our quality assurance process for this dataset, we also examine potential harmful biases and aggressive languages in responses and remove them in the final dataset. We also acknowledge that the generated responses from our experimented models might contain biases.

% \section*{Acknowledgments}
% We thank anonymous reviewers for providing insightful feedback and members from INK and JAUNTS lab. Pei Zhou, Jay Pujara, and Xiang Ren’s work on this project was funded by the Defense Advanced Research Projects Agency with award N660011924033. The research was also supported by gifts from Google.

\section{Limitations}
Our first limitation in modeling CG is that we are using inferences from one speaker to approximate CG during the communication process. To truly represent CG, we need to recollect dialogues and as participants continue the conversations, we should ask both of them the same inference questions and perform post-hoc analysis on the answers to the questions.

Our second limitation is the lack of explicitly modeling \emph{communicative intents}. In future work, we plan to heuristically link each inference dimension to a general communication goal. For example, making inferences about ``speaker emotion states'' is helpful to build emotional connections with the other speaker.\label{limitations}

% Entries for the entire Anthology, followed by custom entries
\bibliographystyle{acl_natbib}
\bibliography{custom}

\begin{thebibliography}{36}
\expandafter\ifx\csname natexlab\endcsname\relax\def\natexlab#1{#1}\fi

\bibitem[{Adolphs et~al.(2021)Adolphs, Shuster, Urbanek, Szlam, and
  Weston}]{adolphs2021reason}
Leonard Adolphs, Kurt Shuster, Jack Urbanek, Arthur Szlam, and Jason Weston.
  2021.
\newblock Reason first, then respond: Modular generation for knowledge-infused
  dialogue.
\newblock \emph{arXiv preprint arXiv:2111.05204}.

\bibitem[{Allwood(1976)}]{allwood1976linguistic}
Jens Allwood. 1976.
\newblock \emph{Linguistic communication as action and cooperation}.
\newblock University of G{\"o}teborg. Department of Linguistics.

\bibitem[{Bohm et~al.(2004)Bohm, Senge, and Nichol}]{bohm2004dialogue}
David Bohm, Peter~M Senge, and Lee Nichol. 2004.
\newblock \emph{On dialogue}.
\newblock Routledge.

\bibitem[{Brown et~al.(2020)Brown, Mann, Ryder, Subbiah, Kaplan, Dhariwal,
  Neelakantan, Shyam, Sastry, Askell et~al.}]{brown2020language}
Tom~B Brown, Benjamin Mann, Nick Ryder, Melanie Subbiah, Jared Kaplan, Prafulla
  Dhariwal, Arvind Neelakantan, Pranav Shyam, Girish Sastry, Amanda Askell,
  et~al. 2020.
\newblock Language models are few-shot learners.
\newblock \emph{arXiv preprint arXiv:2005.14165}.

\bibitem[{Bruner(1961)}]{bruner1961act}
Jerome~S Bruner. 1961.
\newblock The act of discovery.
\newblock \emph{Harvard educational review}.

\bibitem[{Cho and May(2020)}]{cho-may-2020-grounding}
Hyundong Cho and Jonathan May. 2020.
\newblock \href {https://doi.org/10.18653/v1/2020.acl-main.218} {Grounding
  conversations with improvised dialogues}.
\newblock In \emph{Proceedings of the 58th Annual Meeting of the Association
  for Computational Linguistics}, pages 2398--2413, Online. Association for
  Computational Linguistics.

\bibitem[{Clark and Brennan(1991)}]{clark1991grounding}
Herbert~H Clark and Susan~E Brennan. 1991.
\newblock Grounding in communication.

\bibitem[{Clark and Schaefer(1989)}]{clark1989contributing}
Herbert~H Clark and Edward~F Schaefer. 1989.
\newblock Contributing to discourse.
\newblock \emph{Cognitive science}, 13(2):259--294.

\bibitem[{Cui et~al.(2020)Cui, Wu, Liu, Zhang, and Zhou}]{cui2020mutual}
Leyang Cui, Yu~Wu, Shujie Liu, Yue Zhang, and Ming Zhou. 2020.
\newblock \href {https://doi.org/10.18653/v1/2020.acl-main.130} {{M}u{T}ual: A
  dataset for multi-turn dialogue reasoning}.
\newblock In \emph{Proceedings of the 58th Annual Meeting of the Association
  for Computational Linguistics}, pages 1406--1416, Online. Association for
  Computational Linguistics.

\bibitem[{Evans(2003)}]{evans2003two}
Jonathan St~BT Evans. 2003.
\newblock In two minds: dual-process accounts of reasoning.
\newblock \emph{Trends in cognitive sciences}, 7(10):454--459.

\bibitem[{Fleiss(1971)}]{fleiss1971measuring}
Joseph~L Fleiss. 1971.
\newblock Measuring nominal scale agreement among many raters.
\newblock \emph{Psychological bulletin}, 76(5):378.

\bibitem[{Ghosal et~al.(2022)Ghosal, Shen, Majumder, Mihalcea, and
  Poria}]{ghosal2022cicero}
Deepanway Ghosal, Siqi Shen, Navonil Majumder, Rada Mihalcea, and Soujanya
  Poria. 2022.
\newblock Cicero: A dataset for contextualized commonsense inference in
  dialogues.
\newblock \emph{arXiv preprint arXiv:2203.13926}.

\bibitem[{Grice(1975)}]{grice1975logic}
Herbert~P Grice. 1975.
\newblock Logic and conversation.
\newblock In \emph{Speech acts}, pages 41--58. Brill.

\bibitem[{Habermas(1985)}]{habermas1985theory}
J{\"u}rgen Habermas. 1985.
\newblock \emph{The theory of communicative action: Volume 1: Reason and the
  rationalization of society}, volume~1.
\newblock Beacon press.

\bibitem[{Holtzman et~al.(2019)Holtzman, Buys, Du, Forbes, and
  Choi}]{holtzman2019curious}
Ari Holtzman, Jan Buys, Li~Du, Maxwell Forbes, and Yejin Choi. 2019.
\newblock The curious case of neural text degeneration.
\newblock In \emph{International Conference on Learning Representations}.

\bibitem[{Hwang et~al.(2021)Hwang, Bhagavatula, Bras, Da, Sakaguchi, Bosselut,
  and Choi}]{Hwang2021COMETATOMIC2O}
Jena~D. Hwang, Chandra Bhagavatula, Ronan~Le Bras, Jeff Da, Keisuke Sakaguchi,
  Antoine Bosselut, and Yejin Choi. 2021.
\newblock Comet-atomic 2020: On symbolic and neural commonsense knowledge
  graphs.
\newblock In \emph{AAAI}.

\bibitem[{Kahneman(2011)}]{kahneman2011thinking}
Daniel Kahneman. 2011.
\newblock \emph{Thinking, fast and slow}.
\newblock Macmillan.

\bibitem[{Li et~al.(2017)Li, Su, Shen, Li, Cao, and Niu}]{li2017dailydialog}
Yanran Li, Hui Su, Xiaoyu Shen, Wenjie Li, Ziqiang Cao, and Shuzi Niu. 2017.
\newblock \href {https://aclanthology.org/I17-1099} {{D}aily{D}ialog: A
  manually labelled multi-turn dialogue dataset}.
\newblock In \emph{Proceedings of the Eighth International Joint Conference on
  Natural Language Processing (Volume 1: Long Papers)}, pages 986--995, Taipei,
  Taiwan. Asian Federation of Natural Language Processing.

\bibitem[{{Miller} et~al.(2017){Miller}, {Feng}, {Fisch}, {Lu}, {Batra},
  {Bordes}, {Parikh}, and {Weston}}]{miller2017parlai}
A.~H. {Miller}, W.~{Feng}, A.~{Fisch}, J.~{Lu}, D.~{Batra}, A.~{Bordes},
  D.~{Parikh}, and J.~{Weston}. 2017.
\newblock Parlai: A dialog research software platform.
\newblock \emph{arXiv preprint arXiv:{1705.06476}}.

\bibitem[{Rashkin et~al.(2019)Rashkin, Smith, Li, and
  Boureau}]{rashkin2019towards}
Hannah Rashkin, Eric~Michael Smith, Margaret Li, and Y-Lan Boureau. 2019.
\newblock \href {https://doi.org/10.18653/v1/P19-1534} {Towards empathetic
  open-domain conversation models: A new benchmark and dataset}.
\newblock In \emph{Proceedings of the 57th Annual Meeting of the Association
  for Computational Linguistics}, pages 5370--5381, Florence, Italy.
  Association for Computational Linguistics.

\bibitem[{Reimers and Gurevych(2019)}]{reimers-2019-sentence-bert}
Nils Reimers and Iryna Gurevych. 2019.
\newblock \href {http://arxiv.org/abs/1908.10084} {Sentence-bert: Sentence
  embeddings using siamese bert-networks}.
\newblock In \emph{Proceedings of the 2019 Conference on Empirical Methods in
  Natural Language Processing}. Association for Computational Linguistics.

\bibitem[{Roller et~al.(2021)Roller, Dinan, Goyal, Ju, Williamson, Liu, Xu,
  Ott, Smith, Boureau, and Weston}]{roller2020recipes}
Stephen Roller, Emily Dinan, Naman Goyal, Da~Ju, Mary Williamson, Yinhan Liu,
  Jing Xu, Myle Ott, Eric~Michael Smith, Y-Lan Boureau, and Jason Weston. 2021.
\newblock \href {https://aclanthology.org/2021.eacl-main.24} {Recipes for
  building an open-domain chatbot}.
\newblock In \emph{Proceedings of the 16th Conference of the European Chapter
  of the Association for Computational Linguistics: Main Volume}, pages
  300--325, Online. Association for Computational Linguistics.

\bibitem[{Sap et~al.(2019{\natexlab{a}})Sap, Bras, Allaway, Bhagavatula,
  Lourie, Rashkin, Roof, Smith, and Choi}]{sap2019atomic}
Maarten Sap, Ronan~Le Bras, Emily Allaway, Chandra Bhagavatula, Nicholas
  Lourie, Hannah Rashkin, Brendan Roof, Noah~A. Smith, and Yejin Choi.
  2019{\natexlab{a}}.
\newblock \href {https://doi.org/10.1609/aaai.v33i01.33013027} {{ATOMIC:} an
  atlas of machine commonsense for if-then reasoning}.
\newblock In \emph{The Thirty-Third {AAAI} Conference on Artificial
  Intelligence, {AAAI} 2019, The Thirty-First Innovative Applications of
  Artificial Intelligence Conference, {IAAI} 2019, The Ninth {AAAI} Symposium
  on Educational Advances in Artificial Intelligence, {EAAI} 2019, Honolulu,
  Hawaii, USA, January 27 - February 1, 2019}, pages 3027--3035. {AAAI} Press.

\bibitem[{Sap et~al.(2019{\natexlab{b}})Sap, Rashkin, Chen, Le~Bras, and
  Choi}]{sap2019social}
Maarten Sap, Hannah Rashkin, Derek Chen, Ronan Le~Bras, and Yejin Choi.
  2019{\natexlab{b}}.
\newblock \href {https://doi.org/10.18653/v1/D19-1454} {Social {IQ}a:
  Commonsense reasoning about social interactions}.
\newblock In \emph{Proceedings of the 2019 Conference on Empirical Methods in
  Natural Language Processing and the 9th International Joint Conference on
  Natural Language Processing (EMNLP-IJCNLP)}, pages 4463--4473, Hong Kong,
  China. Association for Computational Linguistics.

\bibitem[{Serban et~al.(2017)Serban, Sordoni, Lowe, Charlin, Pineau, Courville,
  and Bengio}]{serban2017hierarchical}
Iulian Serban, Alessandro Sordoni, Ryan Lowe, Laurent Charlin, Joelle Pineau,
  Aaron Courville, and Yoshua Bengio. 2017.
\newblock A hierarchical latent variable encoder-decoder model for generating
  dialogues.
\newblock In \emph{Proceedings of the AAAI Conference on Artificial
  Intelligence}, volume~31.

\bibitem[{Shuster et~al.(2022)Shuster, Komeili, Adolphs, Roller, Szlam, and
  Weston}]{shuster2022language}
Kurt Shuster, Mojtaba Komeili, Leonard Adolphs, Stephen Roller, Arthur Szlam,
  and Jason Weston. 2022.
\newblock Language models that seek for knowledge: Modular search \& generation
  for dialogue and prompt completion.
\newblock \emph{arXiv preprint arXiv:2203.13224}.

\bibitem[{Shwartz et~al.(2020)Shwartz, West, Le~Bras, Bhagavatula, and
  Choi}]{shwartz2020unsupervised}
Vered Shwartz, Peter West, Ronan Le~Bras, Chandra Bhagavatula, and Yejin Choi.
  2020.
\newblock Unsupervised commonsense question answering with self-talk.
\newblock In \emph{Proceedings of the 2020 Conference on Empirical Methods in
  Natural Language Processing (EMNLP)}, pages 4615--4629.

\bibitem[{Speer et~al.(2017)Speer, Chin, and Havasi}]{speer2017conceptnet}
Robyn Speer, Joshua Chin, and Catherine Havasi. 2017.
\newblock Conceptnet 5.5: An open multilingual graph of general knowledge.
\newblock In \emph{Thirty-first AAAI conference on artificial intelligence}.

\bibitem[{Stalnaker(1978)}]{stalnaker1978assertion}
Robert~C Stalnaker. 1978.
\newblock Assertion.
\newblock In \emph{Pragmatics}, pages 315--332. Brill.

\bibitem[{Stanovich and West(2000)}]{stanovich2000individual}
Keith~E Stanovich and Richard~F West. 2000.
\newblock Individual differences in reasoning: Implications for the rationality
  debate?
\newblock \emph{Behavioral and brain sciences}, 23(5):645--665.

\bibitem[{Thoppilan et~al.(2022)Thoppilan, De~Freitas, Hall, Shazeer,
  Kulshreshtha, Cheng, Jin, Bos, Baker, Du et~al.}]{thoppilan2022lamda}
Romal Thoppilan, Daniel De~Freitas, Jamie Hall, Noam Shazeer, Apoorv
  Kulshreshtha, Heng-Tze Cheng, Alicia Jin, Taylor Bos, Leslie Baker, Yu~Du,
  et~al. 2022.
\newblock Lamda: Language models for dialog applications.
\newblock \emph{arXiv preprint arXiv:2201.08239}.

\bibitem[{Wells(2000)}]{wells2000dialogic}
Gordon Wells. 2000.
\newblock Dialogic inquiry in education.
\newblock \emph{Vygotskian perspectives on literacy research}, pages 51--85.

\bibitem[{Zhang et~al.(2020)Zhang, Sun, Galley, Chen, Brockett, Gao, Gao, Liu,
  and Dolan}]{zhang-etal-2020-dialogpt}
Yizhe Zhang, Siqi Sun, Michel Galley, Yen-Chun Chen, Chris Brockett, Xiang Gao,
  Jianfeng Gao, Jingjing Liu, and Bill Dolan. 2020.
\newblock \href {https://doi.org/10.18653/v1/2020.acl-demos.30} {{DIALOGPT} :
  Large-scale generative pre-training for conversational response generation}.
\newblock In \emph{Proceedings of the 58th Annual Meeting of the Association
  for Computational Linguistics: System Demonstrations}, pages 270--278,
  Online. Association for Computational Linguistics.

\bibitem[{Zhao et~al.(2017)Zhao, Zhao, and Eskenazi}]{zhao2017learning}
Tiancheng Zhao, Ran Zhao, and Maxine Eskenazi. 2017.
\newblock Learning discourse-level diversity for neural dialog models using
  conditional variational autoencoders.
\newblock In \emph{Proceedings of the 55th Annual Meeting of the Association
  for Computational Linguistics (Volume 1: Long Papers)}, pages 654--664.

\bibitem[{Zhou et~al.(2021)Zhou, Gopalakrishnan, Hedayatnia, Kim, Pujara, Ren,
  Liu, and Hakkani-Tur}]{zhou-etal-2021-commonsense}
Pei Zhou, Karthik Gopalakrishnan, Behnam Hedayatnia, Seokhwan Kim, Jay Pujara,
  Xiang Ren, Yang Liu, and Dilek Hakkani-Tur. 2021.
\newblock \href {https://aclanthology.org/2021.sigdial-1.13}
  {Commonsense-focused dialogues for response generation: An empirical study}.
\newblock In \emph{Proceedings of the 22nd Annual Meeting of the Special
  Interest Group on Discourse and Dialogue}, pages 121--132, Singapore and
  Online. Association for Computational Linguistics.

\bibitem[{Zhou et~al.(2022)Zhou, Gopalakrishnan, Hedayatnia, Kim, Pujara, Ren,
  Liu, and Hakkani-Tur}]{zhou2022TBS}
Pei Zhou, Karthik Gopalakrishnan, Behnam Hedayatnia, Seokhwan Kim, Jay Pujara,
  Xiang Ren, Yang Liu, and Dilek Hakkani-Tur. 2022.
\newblock Think before you speak: Using self-talk to generate implicit
  commonsense knowledge for response generation.

\end{thebibliography}

%\appendix
\clearpage
\appendix
\label{appendix}

\section{Data Collection Details}\label{appendix_data}
We engage in active discussions with them in the TurkerNation\footnote{\url{https://www.reddit.com/r/TurkerNation/}} Slack channel and provide detailed feedback after multiple rounds of pilot study to ensure the data quality.

\subsection{Inference Collection} 
\begin{figure*}[tb]
	\centering
	\includegraphics[width=1.0\linewidth]{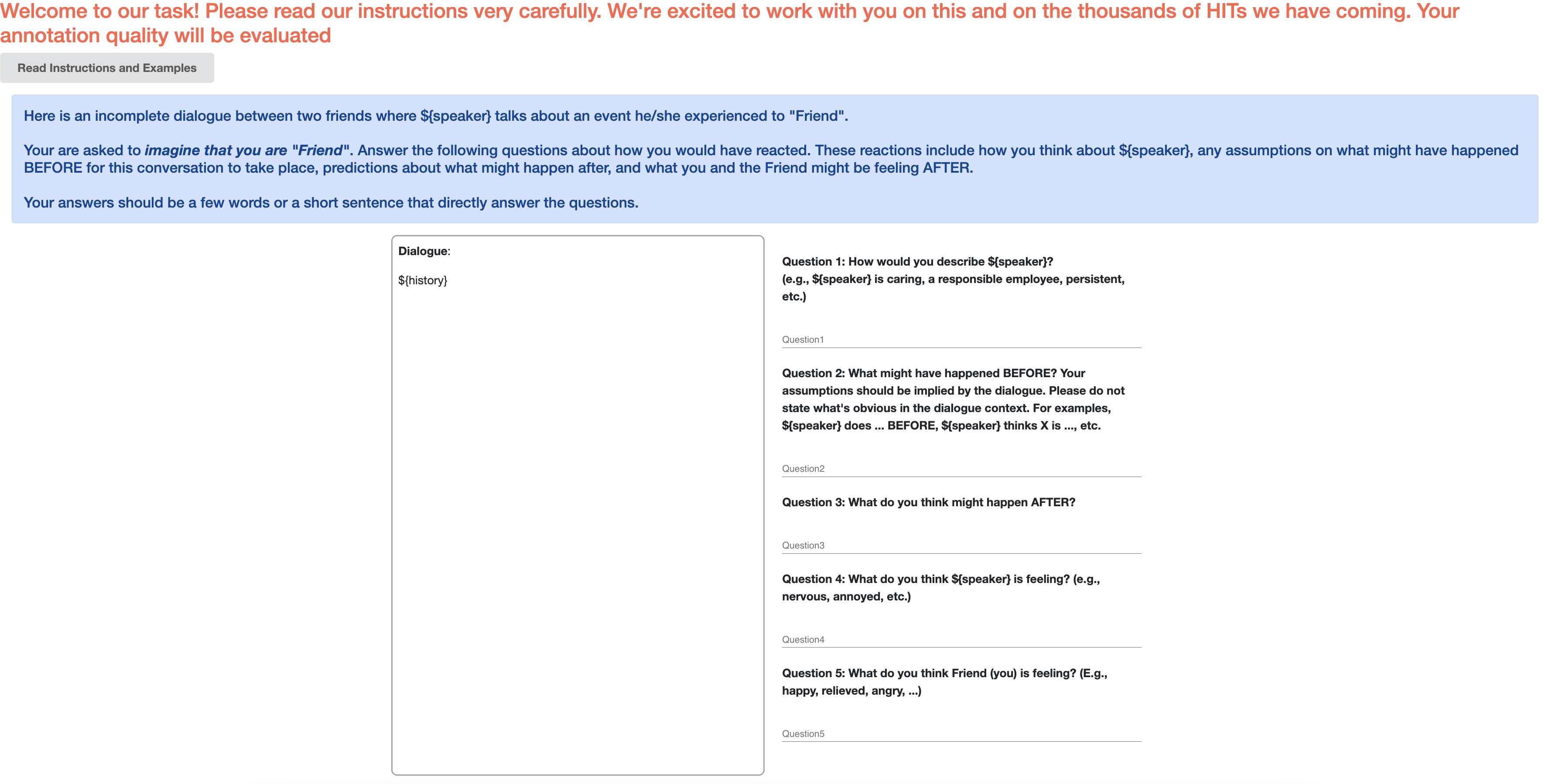}
	\caption{
	\textbf{Inference collection UI}.
	%\xiang{1) Update the figure per our discussion; 2) can be made to double column; flatten the figure (less height) and make each bar thicker}.
	}
	\vspace{-0.1cm}
	\label{fig:inf_col}
\end{figure*}

\begin{figure*}[tb]
	\centering
	\includegraphics[width=1.0\linewidth]{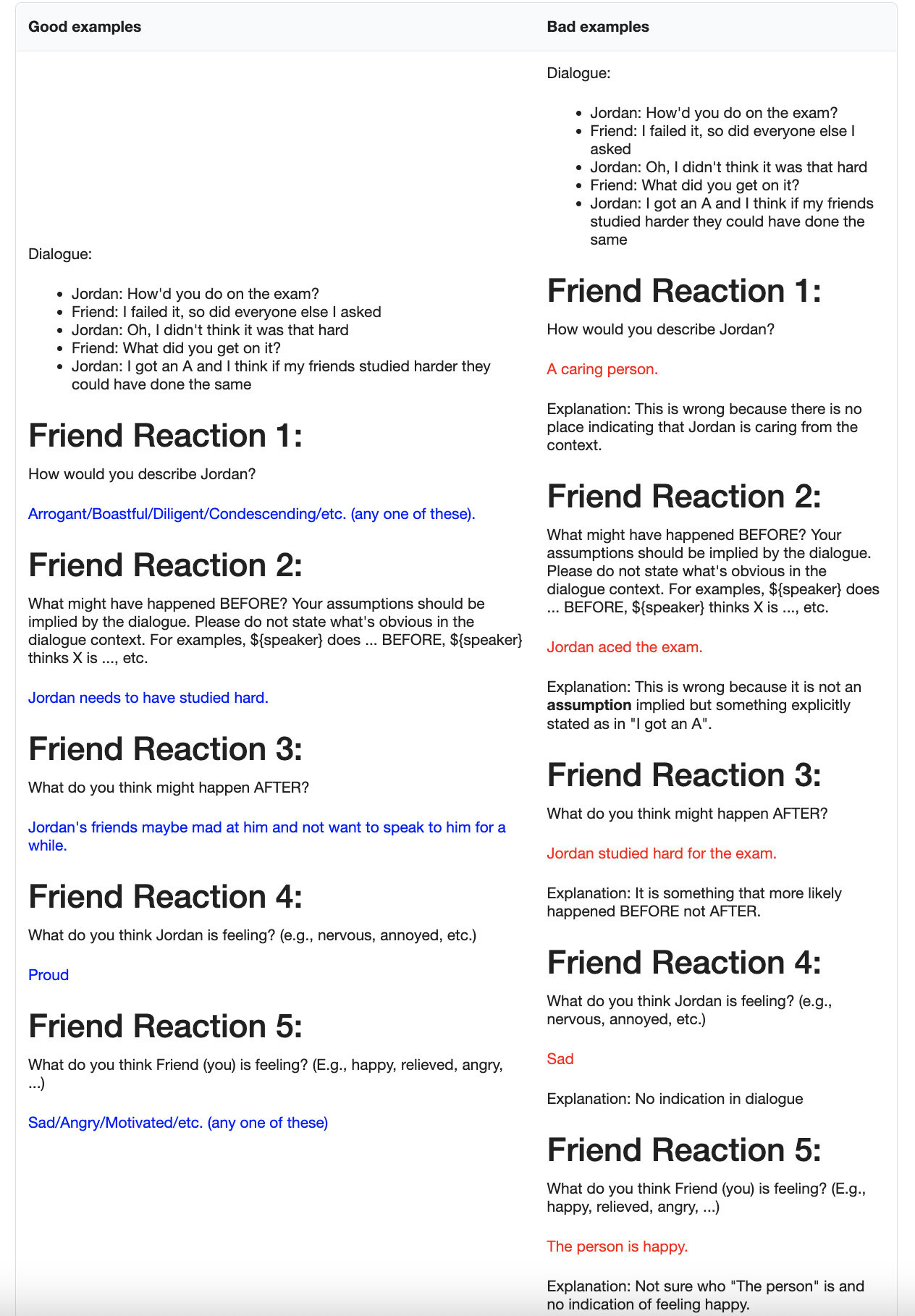}
	\caption{
	\textbf{Inference collection examples for turkers}.
	%\xiang{1) Update the figure per our discussion; 2) can be made to double column; flatten the figure (less height) and make each bar thicker}.
	}
	\vspace{-0.1cm}
	\label{fig:inf_examples}
\end{figure*}

Here we present more detailed feedback for AMT workers on Stage 1. inference collection:
First, we stress that the goal of these answers is to help with generating a response to continue the conversation instead of any inferences that might not be useful for directly generating engaging responses, such as ``\emph{spaghetti is a type of food}'' for the example in Figure~\ref{fig:motivation}.
Secondly, the answers should not be a direct copy-paste of some parts in the dialogue context as those would be trivial to collect, violate the least collaborative principle and the maxim of quantity~\cite{grice1975logic}, and and should not be worth making inferences over.
Finally, we remind them that the inferences written should be considered as ``\emph{common sense}'' so that the approximated CG is more likely to become shared knowledge and beliefs among the dialogue participants. Collection UI and provided examples for turkers are shown in Figures~\ref{fig:inf_col} and~\ref{fig:inf_examples}.

\subsection{Response Collection} 
\begin{figure*}[tb]
	\centering
	\includegraphics[width=1.0\linewidth]{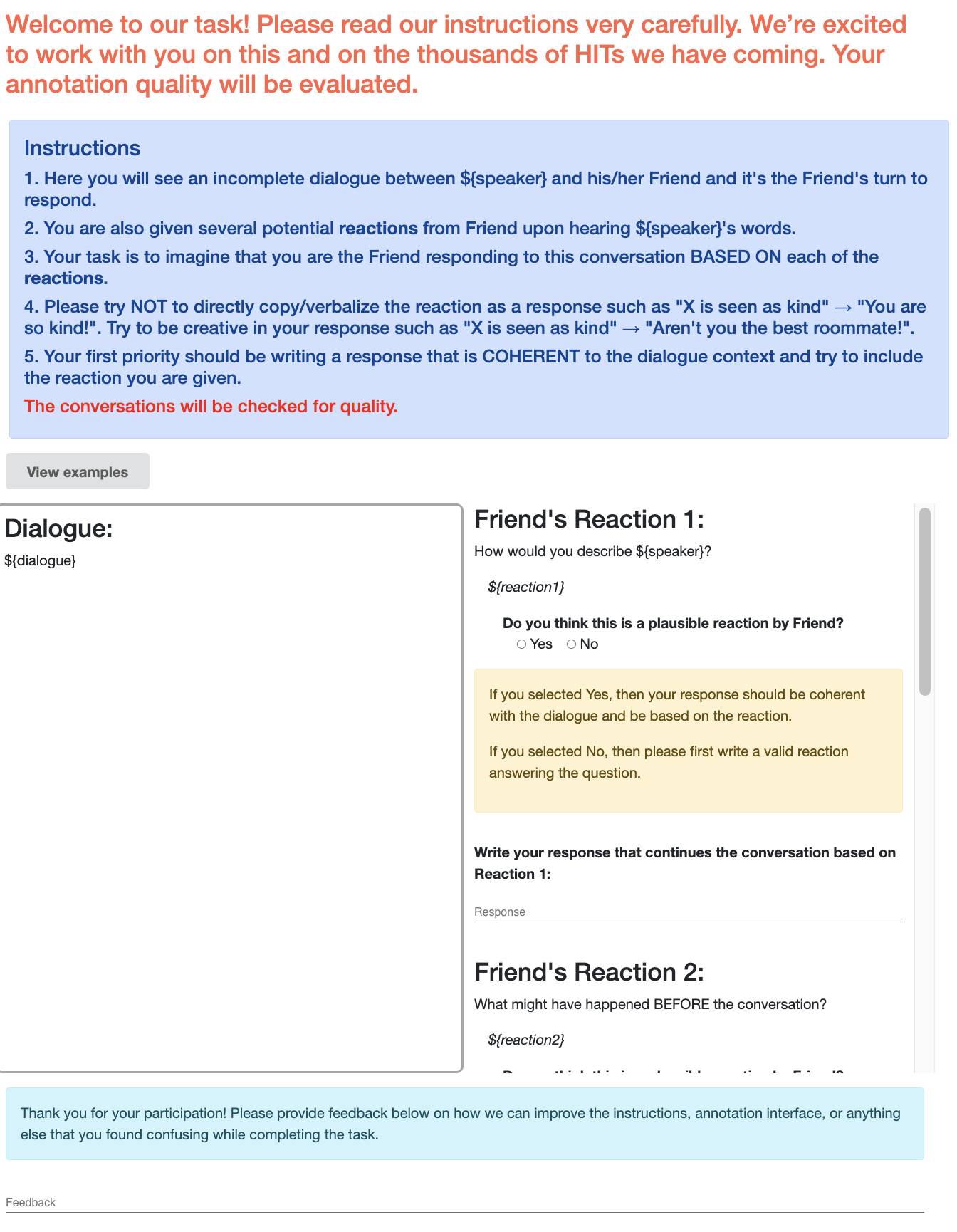}
	\caption{
	\textbf{Response collection UI}.
	%\xiang{1) Update the figure per our discussion; 2) can be made to double column; flatten the figure (less height) and make each bar thicker}.
	}
	\vspace{-0.1cm}
	\label{fig:res_col}
\end{figure*}

\begin{figure*}[tb]
	\centering
	\includegraphics[width=1.0\linewidth]{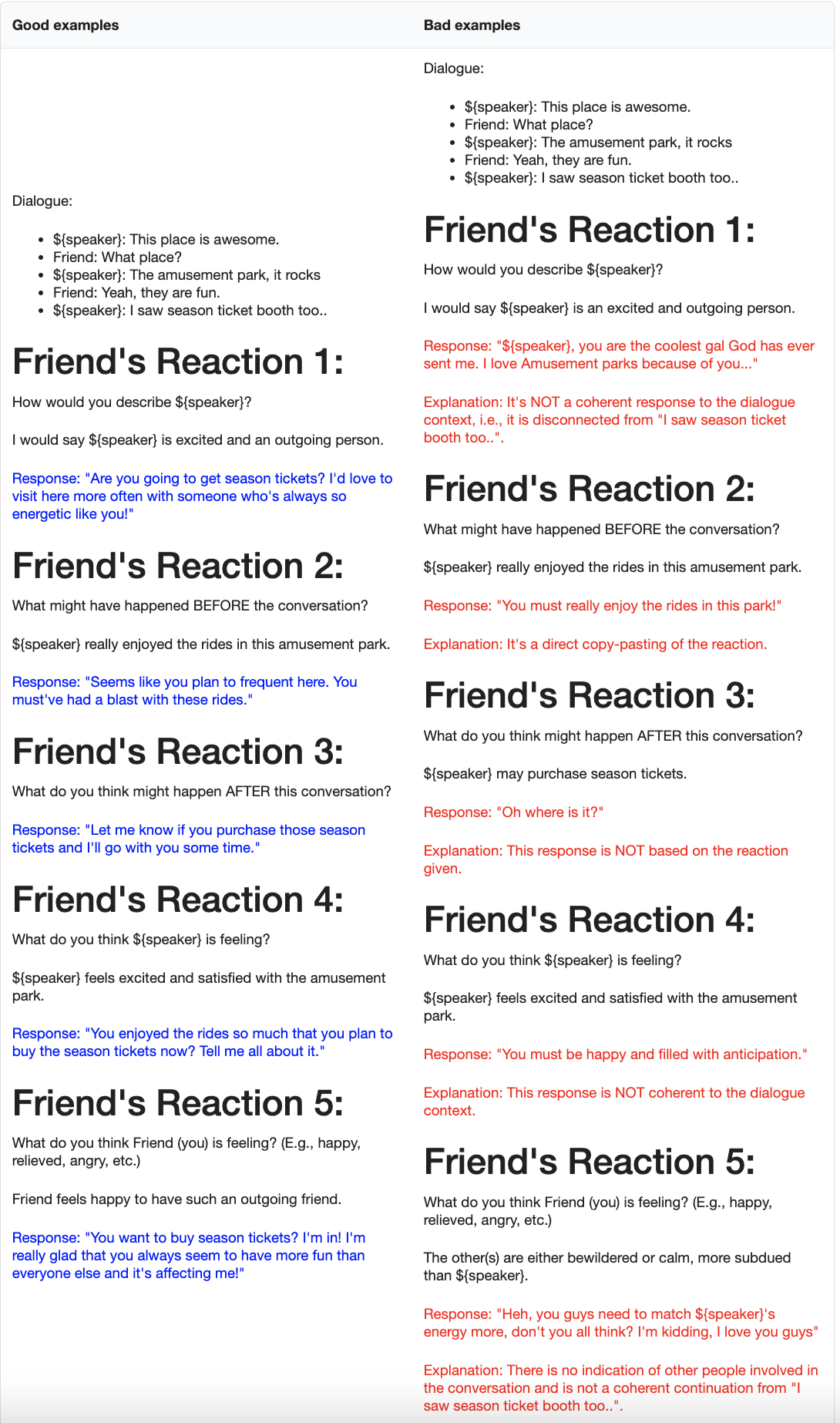}
	\caption{
	\textbf{Response collection examples for turkers}.
	%\xiang{1) Update the figure per our discussion; 2) can be made to double column; flatten the figure (less height) and make each bar thicker}.
	}
	\vspace{-0.1cm}
	\label{fig:res_examples}
\end{figure*}

We specifically stress on several points to workers: 1) to collect more engaging and interesting responses, response should not directly \emph{paraphrase} the inference such as ``\emph{I think you are feeling relieved}'' from inference QA pair ``\emph{What is speaker feeling now? Speaker is feeling relieved}''; 2) the response should be both \emph{coherent} to the dialogue context as what would be naturally uttered by the responder and \emph{based on} the reactions to lead the conversation in an interesting direction; 3) Ultimately, we want responses that lead the conversations that are more enjoyable and engaging.
Collection UI and provided examples for turkers are shown in Figures~\ref{fig:res_col} and~\ref{fig:res_examples}.

\section{Human Evaluation Details}
\begin{figure*}[tb]
	\centering
	\includegraphics[width=1.0\linewidth]{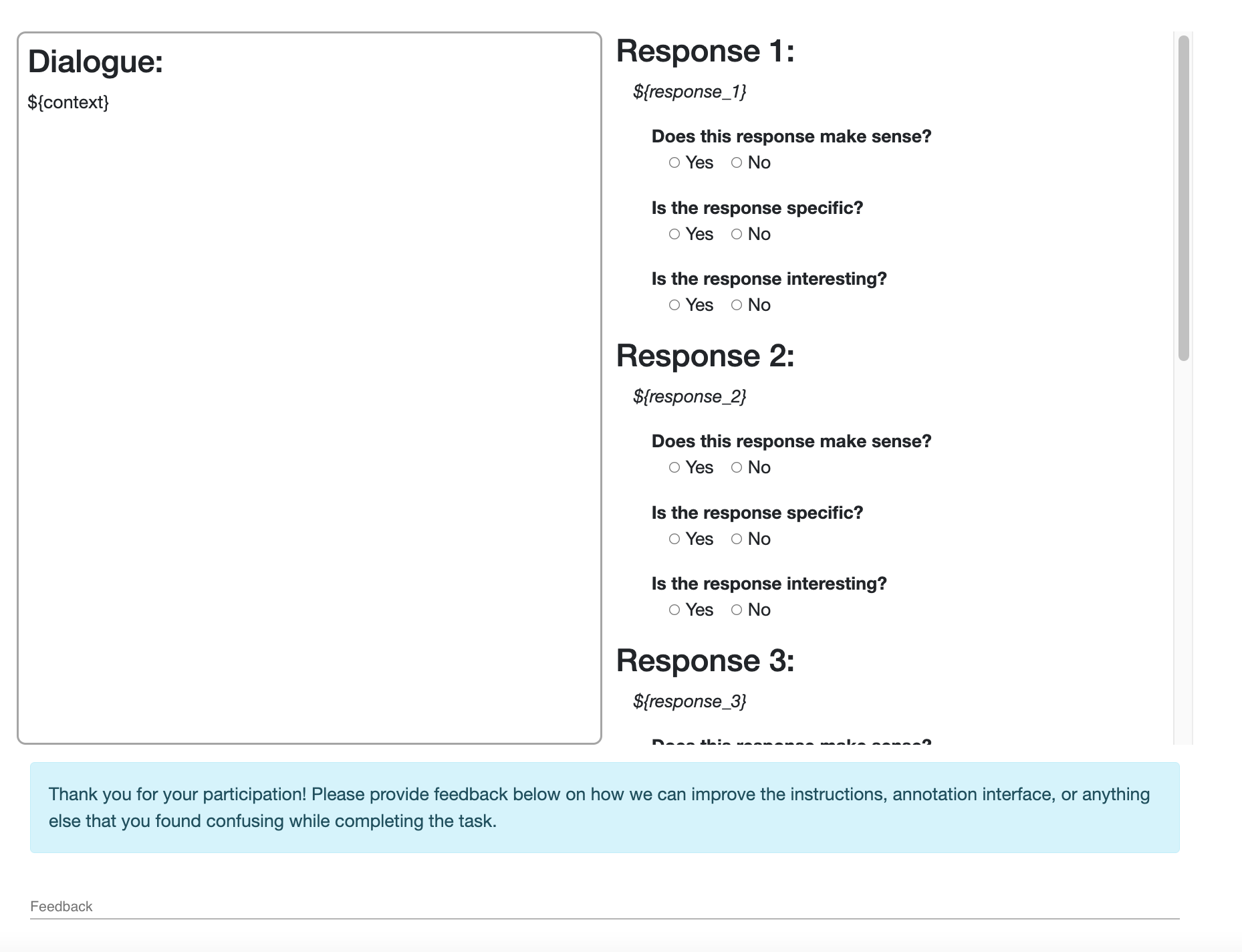}
	\caption{
	\textbf{SSI evaluation UI}.
	%\xiang{1) Update the figure per our discussion; 2) can be made to double column; flatten the figure (less height) and make each bar thicker}.
	}
	\vspace{-0.1cm}
	\label{fig:eval_col}
\end{figure*}

\begin{figure*}[tb]
	\centering
	\includegraphics[width=1.0\linewidth]{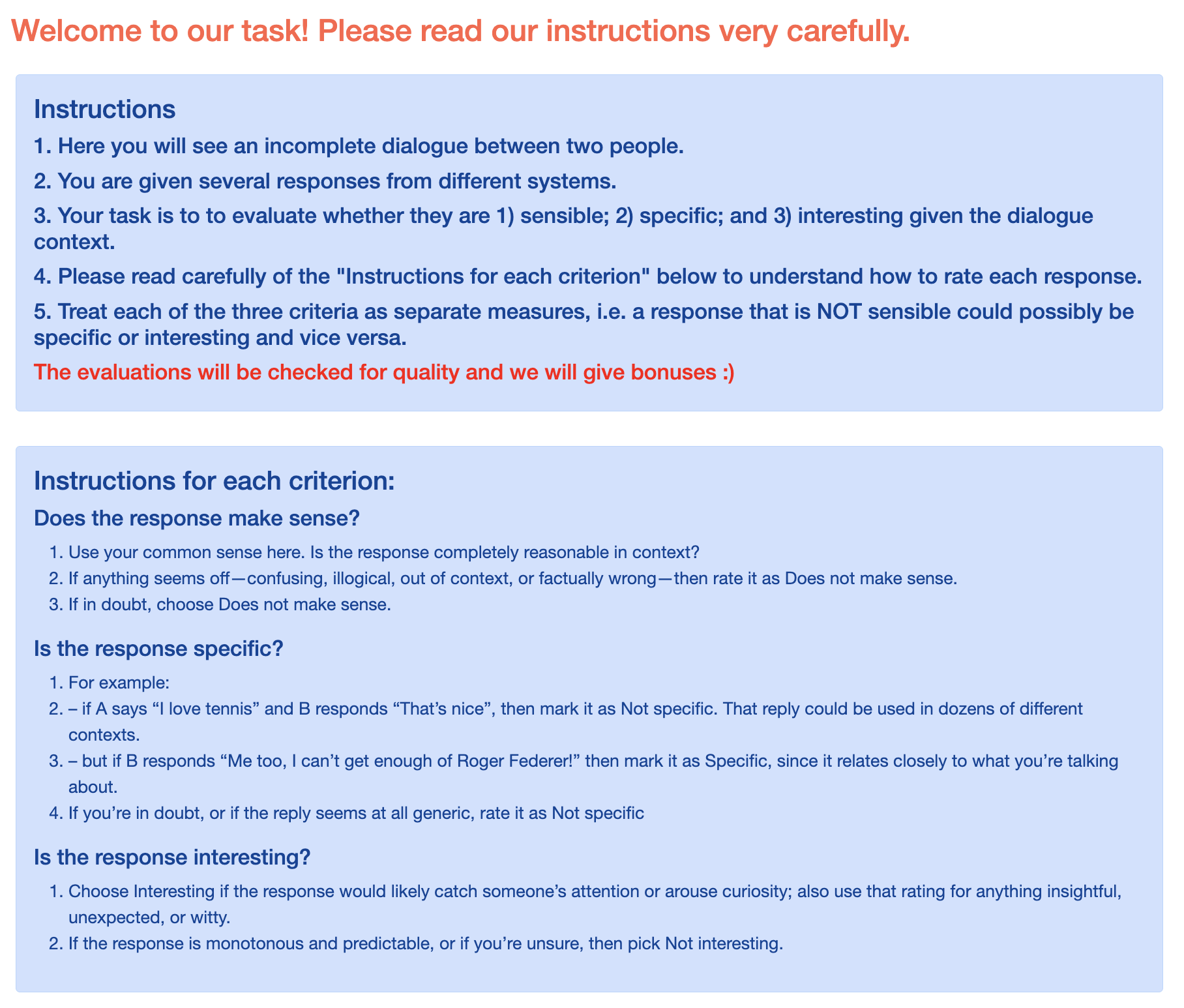}
	\caption{
	\textbf{SSI evaluation instructions}.
	%\xiang{1) Update the figure per our discussion; 2) can be made to double column; flatten the figure (less height) and make each bar thicker}.
	}
	\vspace{-0.1cm}
	\label{fig:eval_ins}
\end{figure*}

Specifically, a sensible response is one that is reasonable in context. A specific response is one that relates closely to the given dialogue context, instead of a generic one that can be applied in dozens of different contexts. An interesting response can ``\emph{catch someone’s attention or arouse their curiosity, or if it is unexpected, witty, or insightful.}''~\cite{thoppilan2022lamda}. For more detailed instructions, please refer to~\citet{thoppilan2022lamda}. Evaluation UI and provided examples for turkers are shown in Figures~\ref{fig:eval_col} and~\ref{fig:eval_ins}.

\section{Model Implementation Details}\label{appendix_implementation}
We use two base models in our paper: BlenderBot-440M and GPT3-175B. For BlenderBot, we use the ParlAI~\cite{miller2017parlai} package for pre-trained modeling and fine-tuning.
The format for fine-tuning BlenderBot on inference questions is: input sequence is ``\emph{<speaker1> ... <speaker2>... <speaker1>... <infq> What might have happened before?}'' and output sequence is ``\emph{<infa>... <speaker2> ...}'', where we use ``\emph{<infq>}'', ``\emph{<infa>}'' to indicate the start of an inference question and answer, respectively.
We fine-tune BlenderBot-440M for 3 epochs with batch size 16 and set the learning rate to be 1e-06. We perform gradient accumulation for 8 steps and gradient clipping with a max norm of 1.0 and optimize using the Adam optimizer. For decoding, we use top-p nucleus sampling~\cite{holtzman2019curious} with temperature T (p = 0.9 and T = 0.7), and a maximum decoding length of 300 tokens. 
BlenderBot-440M models are mostly trained on 4 Quadro RTX 8000 GPUs and take around 9 hours. 

\begin{figure*}[tb]
	\centering
	\includegraphics[width=1.0\linewidth]{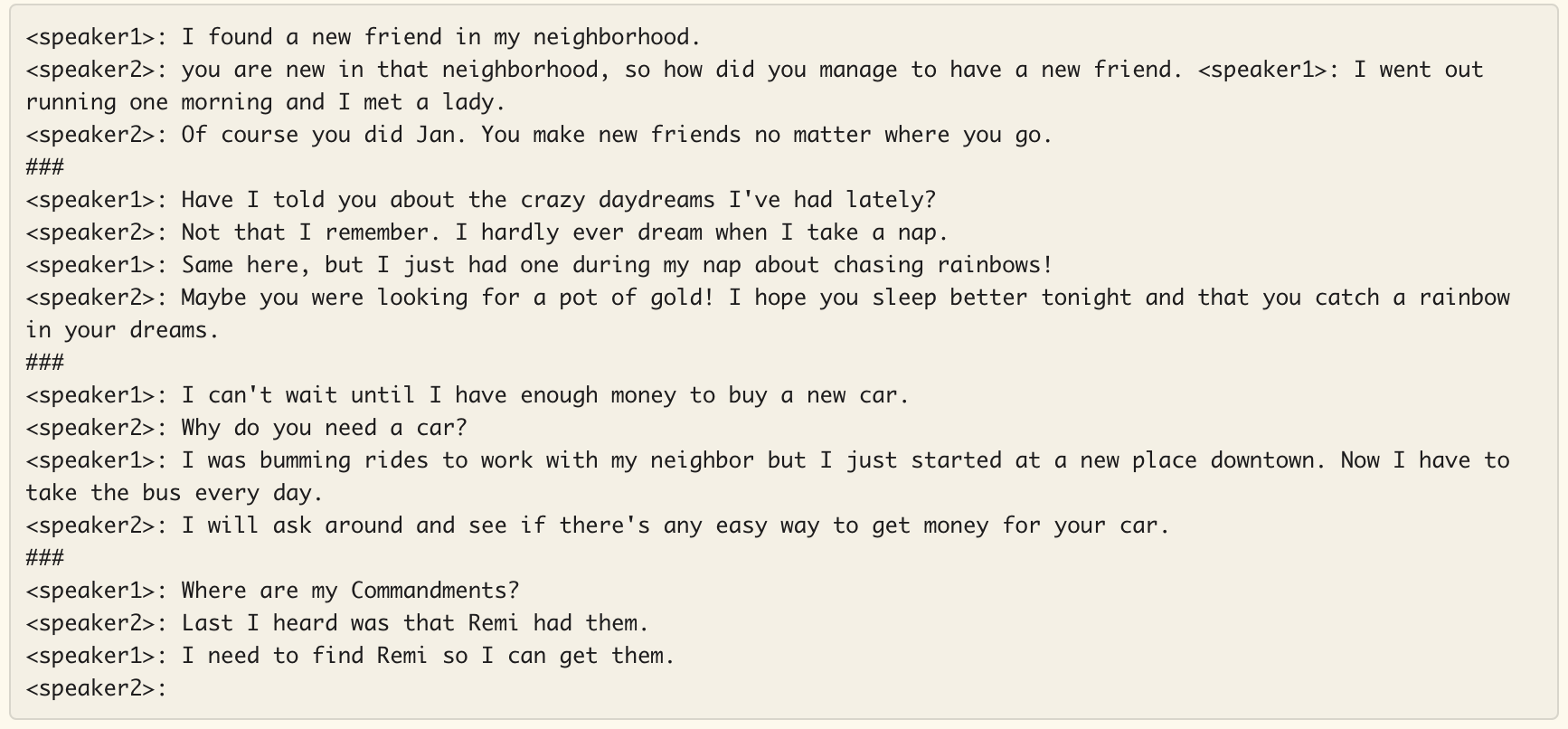}
	\caption{
	\textbf{GPT3-Few Shot Prompting Format (no inference)}.
	%\xiang{1) Update the figure per our discussion; 2) can be made to double column; flatten the figure (less height) and make each bar thicker}.
	}
	\vspace{-0.1cm}
	\label{fig:GPT3_vanilla}
\end{figure*}

\begin{figure*}[tb]
	\centering
	\includegraphics[width=1.0\linewidth]{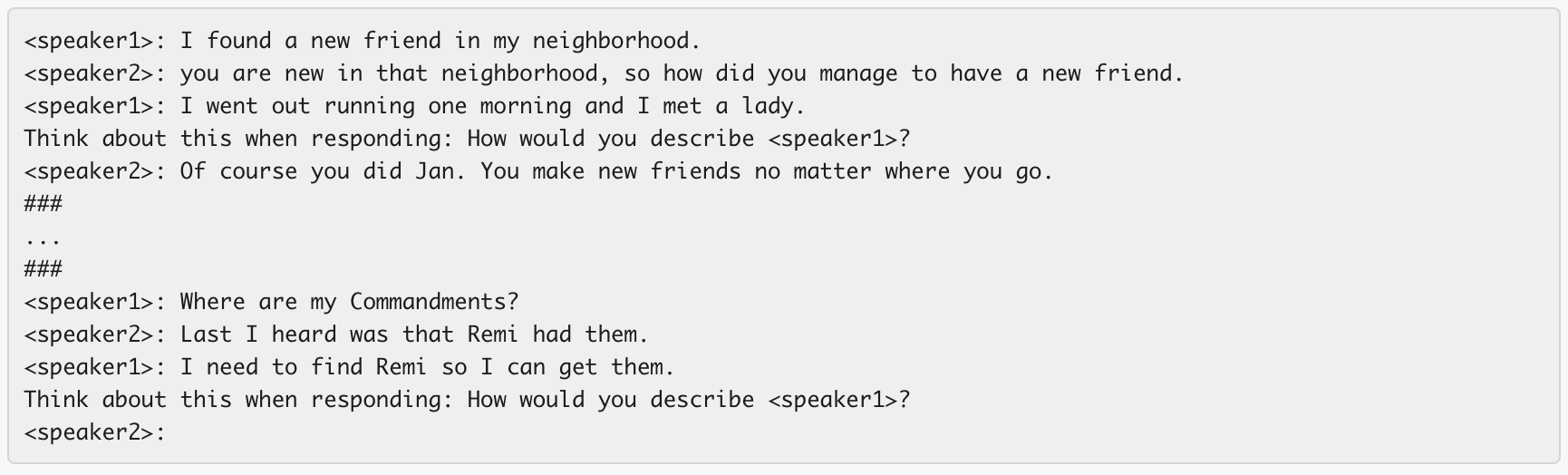}
	\caption{
	\textbf{GPT3-Few Shot-Inference Question Prompting Format}.
	%\xiang{1) Update the figure per our discussion; 2) can be made to double column; flatten the figure (less height) and make each bar thicker}.
	}
	\vspace{-0.1cm}
	\label{fig:GPT3_infq}
\end{figure*}

We use OpenAI-API~\footnote{\url{https://beta.openai.com/playground}} to access GPT3-DaVinci (175B) and include prompting formats for GPT3-FS and GPT3-FS-InfQ in Figures~\ref{fig:GPT3_vanilla}
and~\ref{fig:GPT3_infq}, respectively.

\begin{figure*}[tb]
\centering
\begin{subfigure}[b]{0.4\textwidth}
\includegraphics[width=\textwidth,trim=0cm 0cm 0cm 0cm,clip=true]{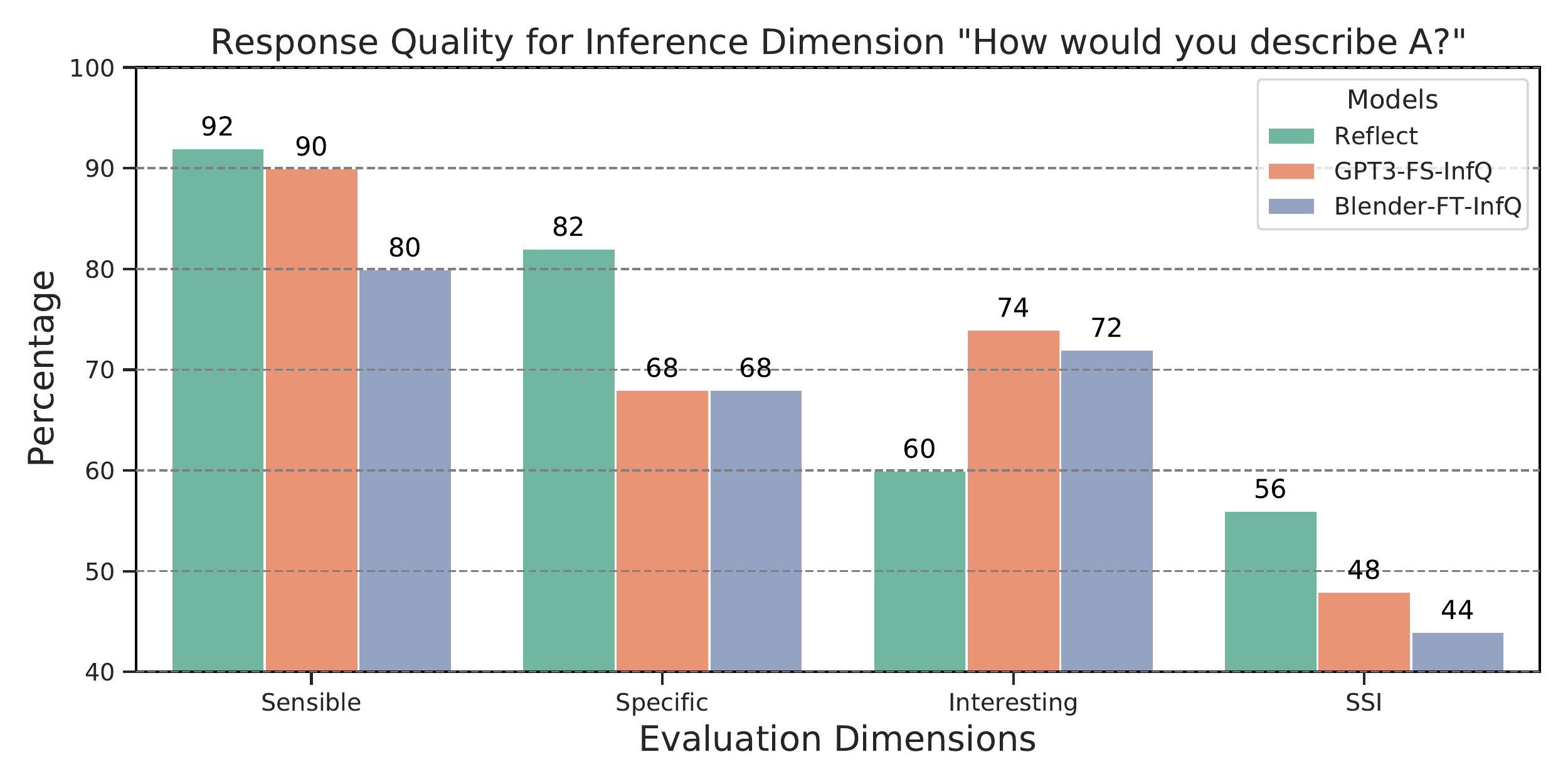}
\end{subfigure}
\begin{subfigure}[b]{0.4\textwidth}
\includegraphics[width=\textwidth,trim=0cm 0cm 0cm 0cm,clip=true]{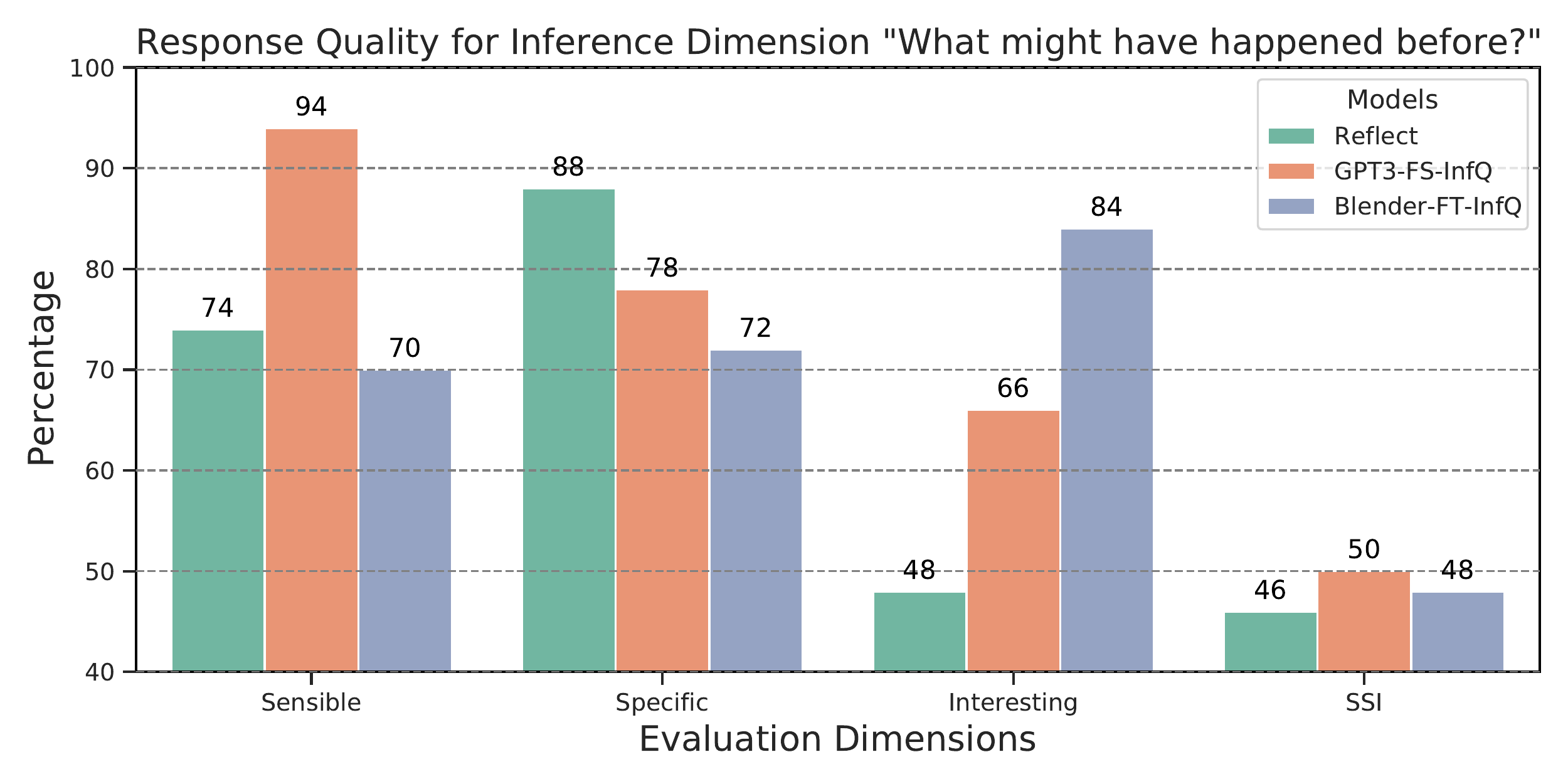}
\end{subfigure}
\begin{subfigure}[b]{0.4\textwidth}
\includegraphics[width=\textwidth]{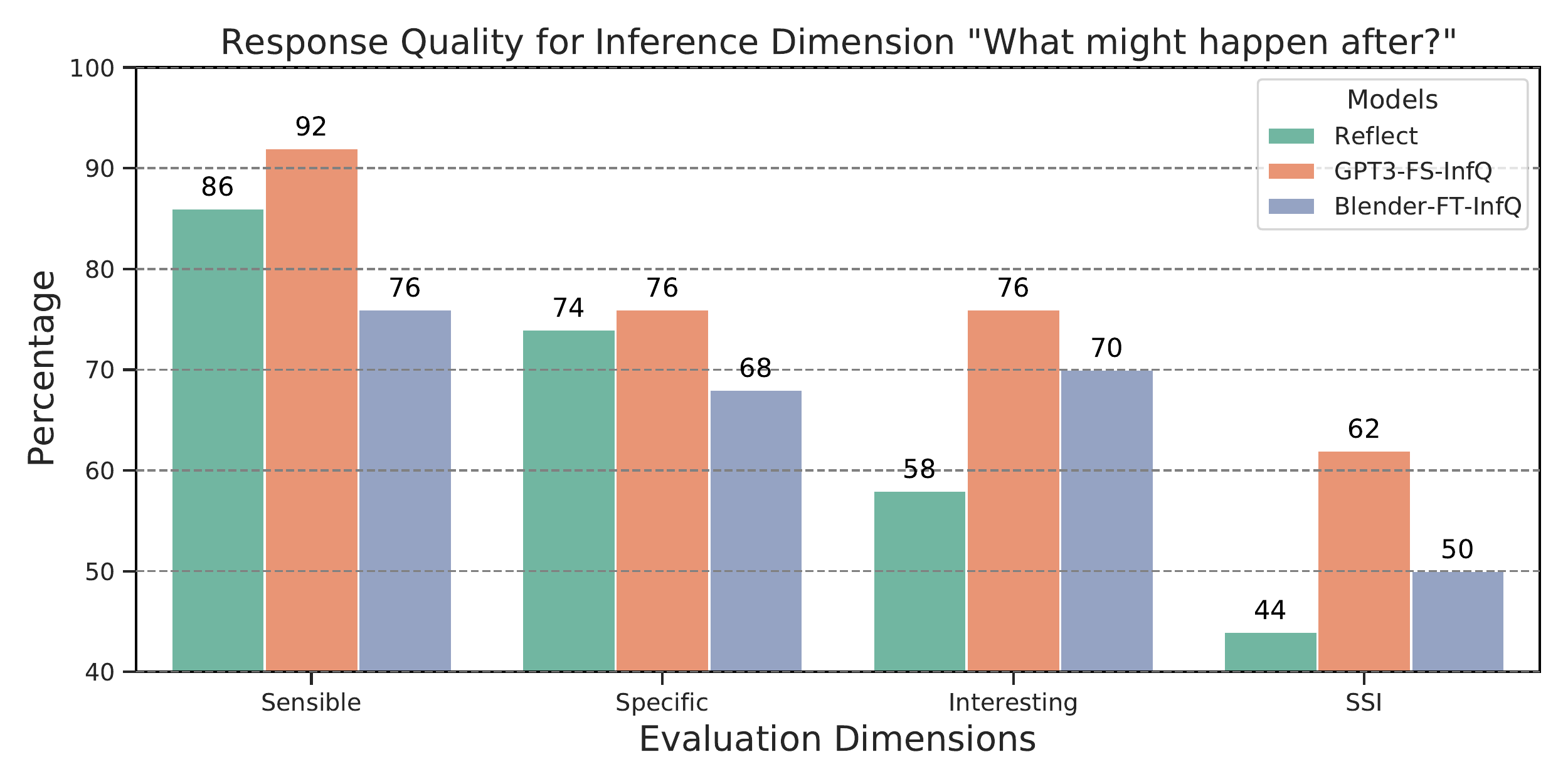}
\end{subfigure}
\begin{subfigure}[b]{0.4\textwidth}
\includegraphics[width=\textwidth]{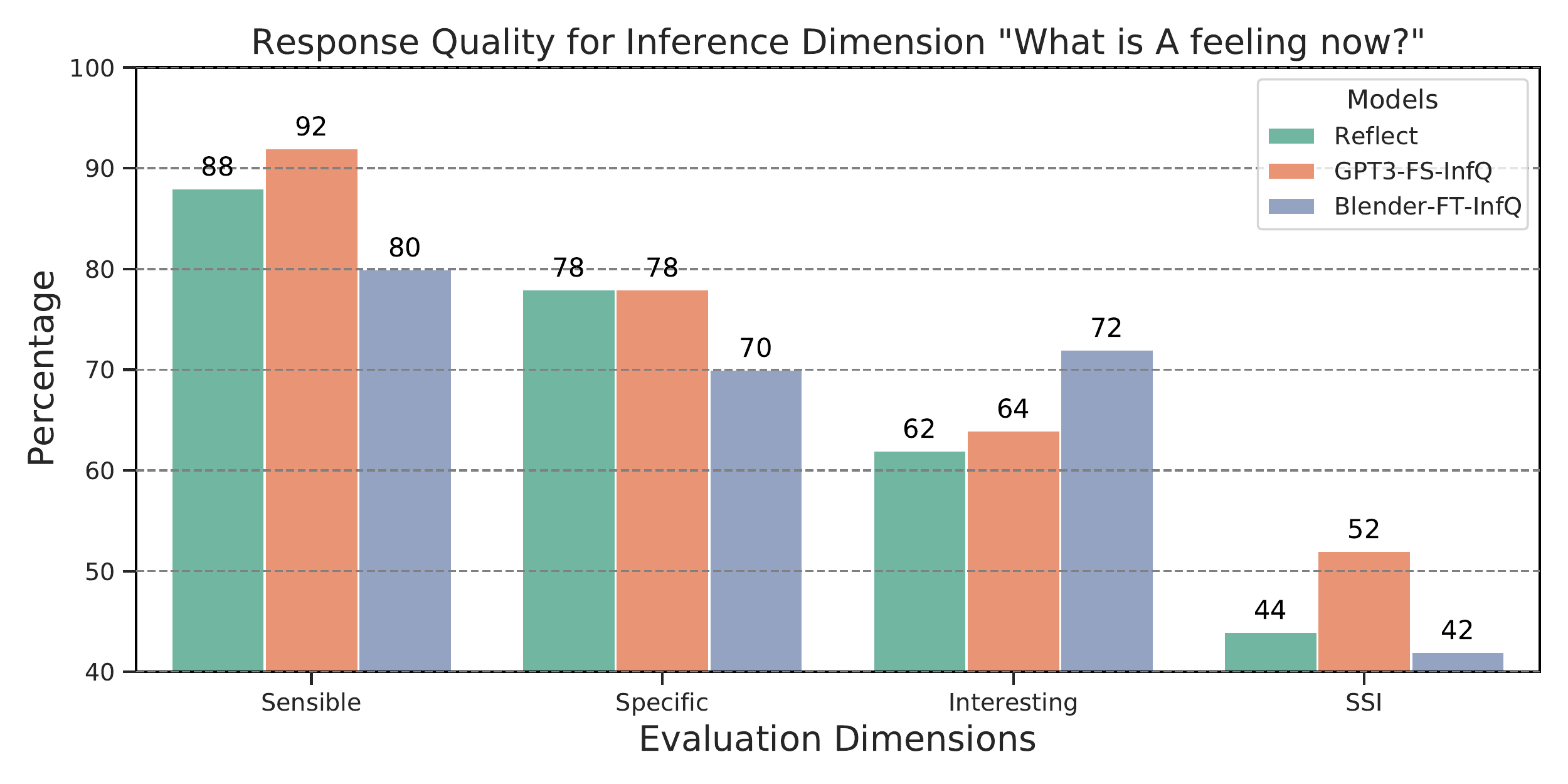}
\end{subfigure}
\begin{subfigure}[b]{0.4\textwidth}
\includegraphics[width=\textwidth]{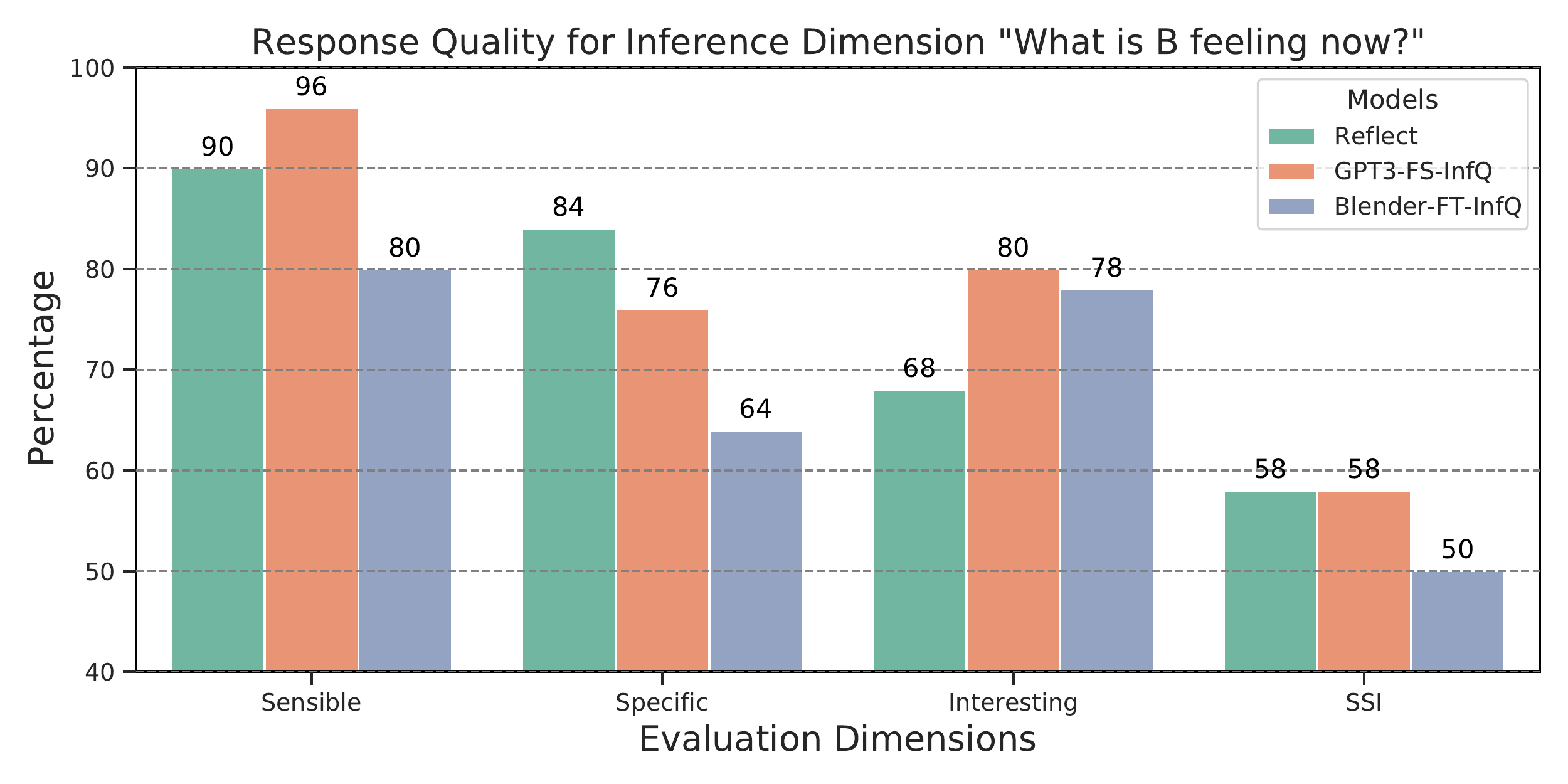}
\end{subfigure}
% \vspace{-0.2cm}
\caption{\small \textbf{Response evaluation separated by inference dimensions}. We find that GPT3-FS-InfQ generate better responses than humans on the potential consequences dimension while generates worse on attributes.
% \xiang{Put all figures into single column space.}
}

\label{fig:inf_separated_appendix}
% \vspace{-0.4cm}
\end{figure*}

\section{Additional Experimental Results}

\subsection{Inference-Separated Fine-Grained Evaluation Results}
Inference dimension-separated full results are shown in Figure~\ref{fig:inf_separated_appendix}.


\end{document}